\documentclass[11pt]{article}

\usepackage[final]{acl}

\usepackage{times}
\usepackage{latexsym}

\usepackage[T1]{fontenc}
\usepackage[utf8]{inputenc}
\usepackage{algorithm}
\usepackage{algpseudocode}

\usepackage{microtype}

\usepackage{inconsolata}

\usepackage{graphicx}

\newcommand\blfootnote[1]{
    \begingroup
    \renewcommand\thefootnote{}\footnote{#1}
    \addtocounter{footnote}{-1}
    \endgroup
}
\usepackage{times}
\usepackage{latexsym}
\usepackage{tcolorbox}
\tcbuselibrary{skins, breakable, theorems}
\usepackage{booktabs}
\usepackage{multirow}
\usepackage{adjustbox}
\usepackage{xcolor}
\usepackage{soul}
\usepackage{graphicx}
\usepackage{subcaption}
\usepackage{caption}
\usepackage{makecell}
\usepackage{algcompatible}
\usepackage{array}
\usepackage{enumitem}

\definecolor{MyBlue}{HTML}{a9d5ee} % a9d5ee
\definecolor{MyLightBlue}{HTML}{DAEEFA}% DAEEFA
\definecolor{DemoOrange}{HTML}{ea7132}
\definecolor{DemoGreen}{HTML}{4ea72d}
\definecolor{DemoBlue}{HTML}{0f9ed5}
\newcommand{\textcolorblue}[1]{
  \begingroup
  \sethlcolor{MyBlue}%背景色
  \textcolor{black}{\hl{#1}}%textcolor里面对应文字颜色
  \endgroup
}
\newcommand{\textcolorlblue}[1]{
  \begingroup
  \sethlcolor{MyLightBlue}%背景色
  \textcolor{black}{\hl{#1}}%textcolor里面对应文字颜色
  \endgroup
}

\title{Mol-Debate: Multi-Agent Debate Improves Structural Reasoning in Molecular Design}

\author{
    \textbf{Wengyu Zhang}$^\spadesuit$ \quad \textbf{Xiao-Yong Wei}$^{\heartsuit, \spadesuit,}$\thanks{\ Corresponding author}\quad \textbf{Qing Li}$^\spadesuit$ \\
    $^\spadesuit$The Hong Kong Polytechnic University \quad $^\heartsuit$Sichuan University\\
    \texttt{weng-yu.zhang@connect.polyu.hk}\\
    \texttt{\{cs007.wei, qing-prof.li\}@polyu.edu.hk}
}

\begin{document}
\maketitle

% %%%%%%%%%%%%%%%%%%%%%%%%%%%%%%%%%%%%%%%%%%%%%%%%%%%%%%%%%%%%%%%%%%%%%%%%%%%%%%%%%%%%%
% %%%%%%%%%%%%%%%%%%%%%%%%%%%%%%%%%%%%%%%%%%%%%%%%%%%%%%%%%%%%%%%%%%%%%%%%%%%%%%%%%%%%%
% %%%%%%%%%%%%%%%%%%%%%%%%%%%%%%%%%%%%%%%%%%%%%%%%%%%%%%%%%%%%%%%%%%%%%%%%%%%%%%%%%%%%%
%
% Abstract
%
% >>>>>>>>>>>>>>>>>>>>>>>>>>>>>>>>>>>>>>>>>>>>>>>>>>>>>>>>>>>>>>>>>>>>>>>>>>>>>>>>>>>>>

\begin{abstract}

Text-guided molecular design is a key capability for AI-driven drug discovery,
yet it remains challenging to map sequential natural-language instructions with non-linear molecular structures under strict chemical constraints.
Most existing approaches, including RAG, CoT prompting, and fine-tuning or RL, emphasize a small set of ad-hoc reasoning perspectives implemented in a largely one-shot generation pipeline.
In contrast, real-world drug discovery relies on dynamic, multi-perspective critique and iterative refinement to reconcile semantic intent with structural feasibility.
Motivated by this, we propose \textbf{Mol-Debate}, a generation paradigm that enables such dynamic reasoning through an iterative generate-debate-refine loop.
We further characterize key challenges in this paradigm and address them through perspective-oriented orchestration, including developer-debater conflict, global-local structural reasoning, and static-dynamic integration.
Experiments demonstrate that Mol-Debate achieves state-of-the-art performance against strong general and chemical baselines, reaching 59.82\% exact match on ChEBI-20 and 50.52\% weighted success rate on S$^2$-Bench.
Our code is available at \url{https://github.com/wyuzh/Mol-Debate}.
\blfootnote{\textit{Preprint.}}
\end{abstract}

% %%%%%%%%%%%%%%%%%%%%%%%%%%%%%%%%%%%%%%%%%%%%%%%%%%%%%%%%%%%%%%%%%%%%%%%%%%%%%%%%%%%%%
% %%%%%%%%%%%%%%%%%%%%%%%%%%%%%%%%%%%%%%%%%%%%%%%%%%%%%%%%%%%%%%%%%%%%%%%%%%%%%%%%%%%%%
% %%%%%%%%%%%%%%%%%%%%%%%%%%%%%%%%%%%%%%%%%%%%%%%%%%%%%%%%%%%%%%%%%%%%%%%%%%%%%%%%%%%%%
%
% Introduction
%
% >>>>>>>>>>>>>>>>>>>>>>>>>>>>>>>>>>>>>>>>>>>>>>>>>>>>>>>>>>>>>>>>>>>>>>>>>>>>>>>>>>>>>

\section{Introduction}
\begin{figure}[t]
    \centering
    \includegraphics[width=1\linewidth]{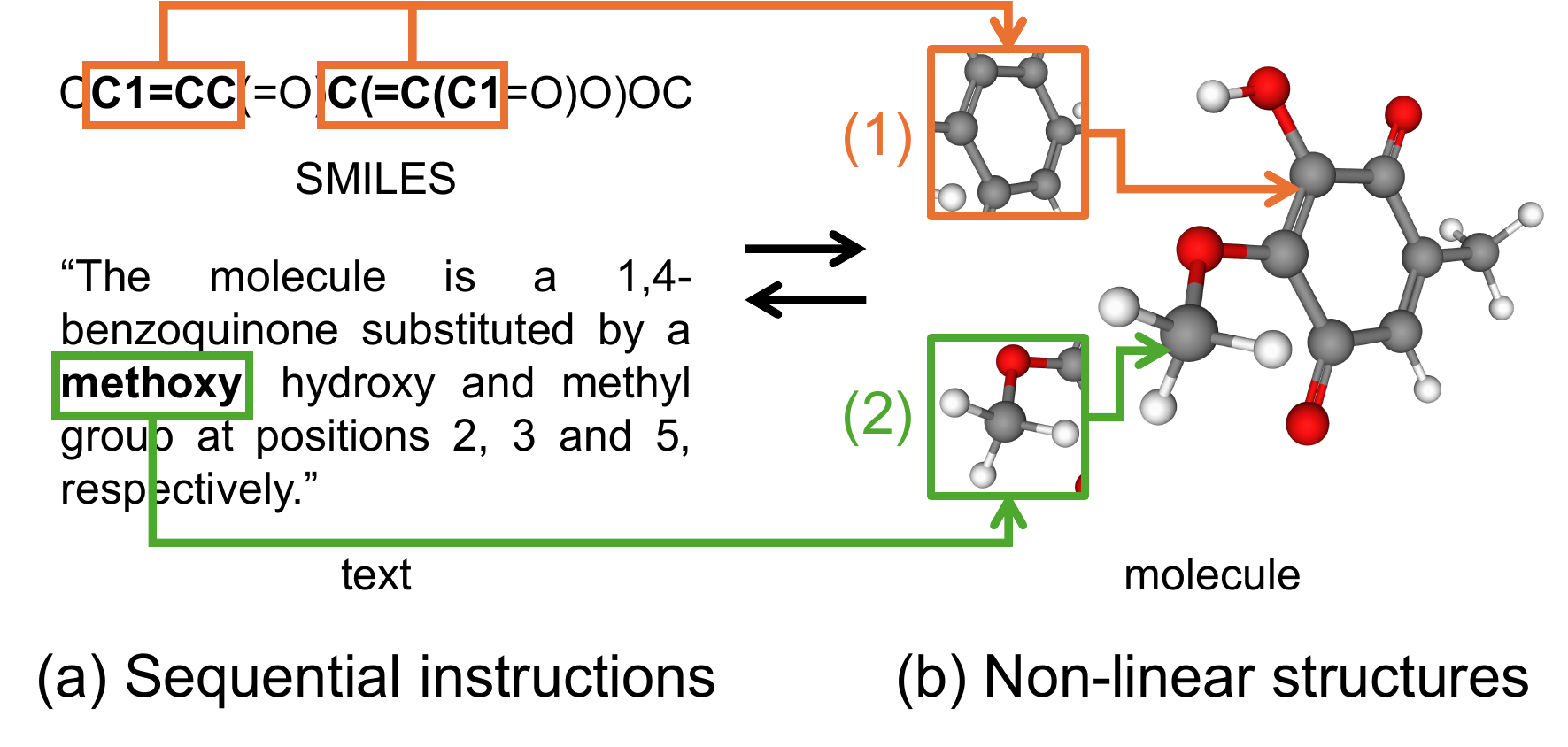}
    \caption{
    Illustration of the text-structure gap: (a) sequential human instructions must be grounded in (b) non-linear structural domain of molecules.
    (1) The ring substructure is split across non-adjacent spans in the SMILES sequence.
    (2) The methoxy substructure appears as a single word in the text but corresponds to multiple atoms and bonds in the molecule.
    }
    
    \label{fig:gap_demo}
\end{figure}

Drug discovery is a focal point of AI for science, posing a critical multimodal NLP challenge despite the broad success of large language models (LLMs)~\cite{edwards2022translation,cao2022identifying,hurst2024gpt,graphatc}.
As shown in Figure~\ref{fig:gap_demo}, the fundamental obstacle is a text-structure gap: translating sequential human instructions into the non-linear structural domain of molecules, which is defined by ring topology and functional group connectivity~\cite{jang-etal-2025-structural}.
A range of methods have been proposed to enhance structural awareness in reasoning. These include attention-based alignment between language and molecular features~\cite{vaswani2017attention,schwaller2019molecular}; Retrieval-Augmented Generation (RAG), which supplies relevant external chemical context~\cite{lewis2020retrieval,li2024empowering}; and Chain-of-Thought (CoT) strategies that encourage structured reasoning~\cite{wei2022chain}. 
Domain-specific models trained via fine-tuning (FT) and reinforcement learning (RL) on large chemistry corpora have also been developed to build foundational chemical competence ~\cite{edwards2022translation,pei2023biot5,zhao2025chemdfm}.

While effective, existing methods are often constrained to a few ad-hoc reasoning perspectives.
For example, RAG mainly strengthens external knowledge retrieval, while pretraining and fine-tuning frameworks largely emphasize token alignment and data-driven pattern learning. As a result, these approaches are often tied to the specific perspective they are designed to exploit.
Real-world drug discovery, however, demands a more integrative approach. 
The progress hinges on chemists’ ability to iteratively reconcile multiple viewpoints, such as high-level intent, structural constraints, and design trade-offs, through collaborative critique and refinement.
While multi-agent systems (MAS) are a promising avenue for integrating the diverse perspectives required in drug design, their application raises a critical new issue~\cite{guo2024large,swanson2025virtual}. 
The system must navigate a delicate balance: preserving agent independence to avoid groupthink while ensuring enough coordination to reach a coherent decision~\cite{li2024survey}. 
If agents are too independent, their outputs fragment; if coordination is too strict, the system may converge prematurely and miss subtle errors.

\begin{figure*}[t]
    \centering
    \includegraphics[width=1\linewidth]{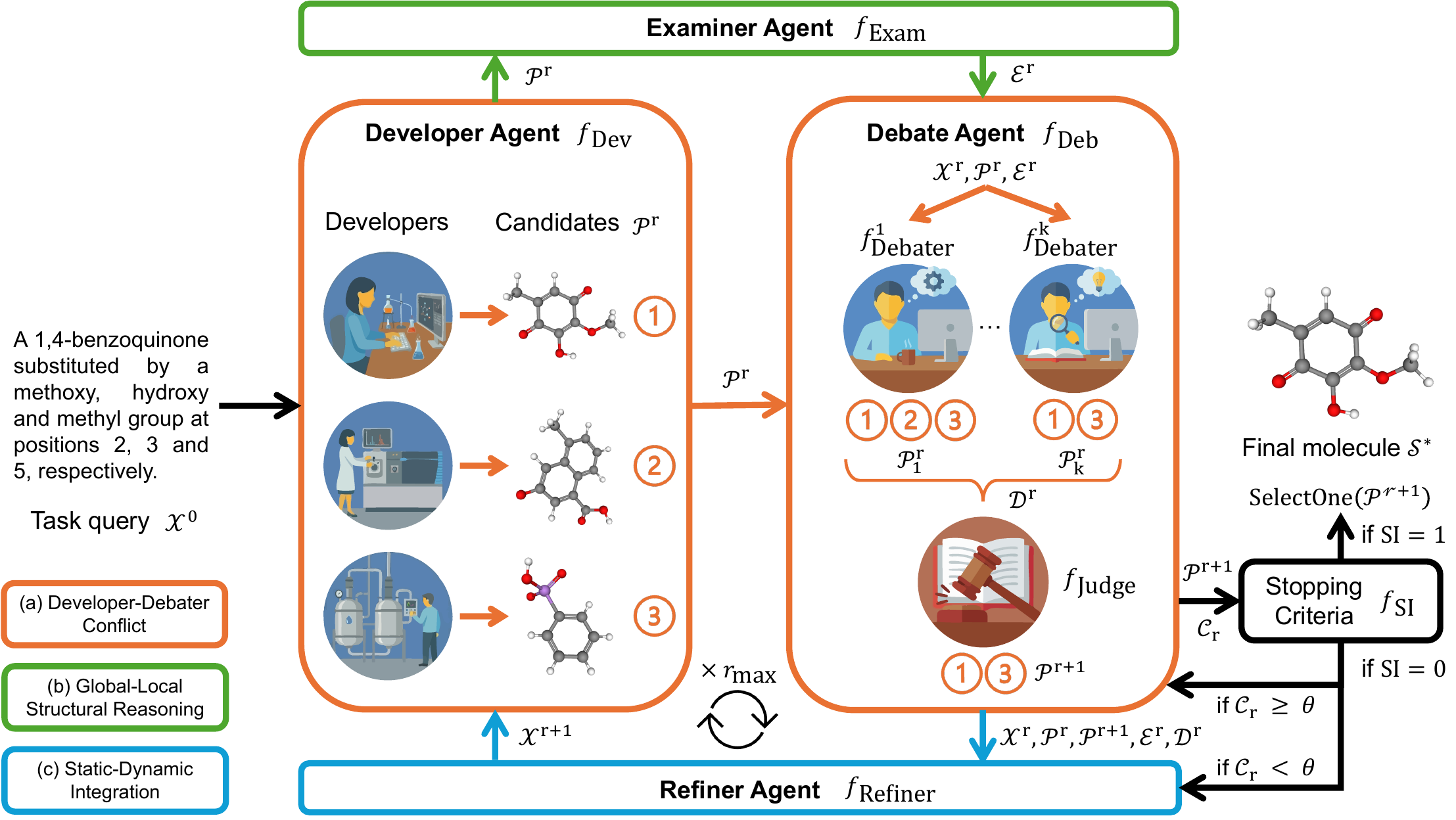}
    \caption{An overview of our Mol-Debate framework, an iterative generation framework where agents collaborate in a generate-debate-refine loop, integrating diverse and dynamic perspectives: (a) Developer-Debater Conflict, (b) Global-Local Structural Reasoning, and (c) Static-Dynamic Integration.
}
    \label{fig:pipeline}
\end{figure*}

This new challenge positions Multi-Agent Debate (MAD) ~\cite{du2023improving} as a compelling, yet unexplored, solution for molecular design, as its core mechanism of debate naturally balances independent reasoning with collaborative coordination. 
Motivated by this, we present a pilot study to investigate if debate-driven agentic reasoning can more effectively bridge the text-structure gap. 
To this end, we introduce \textbf{Mol-Debate}, an iterative generation framework (see Figure~\ref{fig:pipeline}) where agents collaborate in a generate-debate-refine loop, integrating diverse and dynamic perspectives to improve structural fidelity.
Mol-Debate addresses the integration of perspectives at three distinct levels.
\begin{itemize}[leftmargin=*]
    \item \textbf{Developer-Debater Conflict}: A known issue where agents endowed with extensive domain expertise can experience a corresponding limitation in general reasoning and linguistic adaptability for debating~\cite{liu2024more}. To mitigate this asymmetry, we employ dedicated Debater Agents. Their function is to inject common-sense reasoning into the process, critically assessing how well candidate molecules satisfy the semantic intent of the prompt, thereby guaranteeing chemical robustness and precise user alignment.

    \item \textbf{Global-Local Structural Reasoning}: Standard models typically process molecules as global, linear strings, which often causes them to overlook fine-grained chemical rules~\cite{edwards2022translation,dubey2024llama}. Mol-Debate bridges this gap by enforcing a Local Structural Perspective. This perspective integrates a microscopic analysis of chemical constraints, shifting the model's reasoning beyond mere ``textual probability''. Consequently, the system can simultaneously respect both the global molecular syntax and the local chemical reality of its substructures.
    For example, this perspective helps Debater Agents prefer candidates with reasonable molecular weight and a reasonable number of rotatable bonds.

    \item \textbf{Static-Dynamic Integration}: Current pipelines (e.g., RAG~\cite{li2024empowering}, CoT~\cite{jang-etal-2025-structural}, and chemical LLMs~\cite{zhao2025developing}) are fundamentally static, executing a one-way generation where errors are final. In contrast, Mol-Debate establishes a dynamic refinement system. It treats inter-agent disagreement not as failure, but as a learning signal. By creating a closed feedback loop, this consensus dissonance actively reformulates task constraints, allowing the system to refine its understanding of the query during inference, rather than passively accepting the initial output.
\end{itemize}

Our contributions are threefold:
\begin{itemize}
    \item We identify the gap between prior methods that rely on a few ad-hoc reasoning perspectives and real-world drug discovery that requires dynamic, multi-perspective critique and refinement.
    \item We propose \textbf{Mol-Debate}, a generation-by-debate paradigm for molecular design.
    \item We characterize and address key challenges in this paradigm. Mol-Debate achieves the state-of-the-art performance and consistently high chemical validity against strong baselines.
\end{itemize}

% %%%%%%%%%%%%%%%%%%%%%%%%%%%%%%%%%%%%%%%%%%%%%%%%%%%%%%%%%%%%%%%%%%%%%%%%%%%%%%%%%%%%%
% %%%%%%%%%%%%%%%%%%%%%%%%%%%%%%%%%%%%%%%%%%%%%%%%%%%%%%%%%%%%%%%%%%%%%%%%%%%%%%%%%%%%%
% %%%%%%%%%%%%%%%%%%%%%%%%%%%%%%%%%%%%%%%%%%%%%%%%%%%%%%%%%%%%%%%%%%%%%%%%%%%%%%%%%%%%%
%
% Related Work
%
% >>>>>>>>>>>>>>>>>>>>>>>>>>>>>>>>>>>>>>>>>>>>>>>>>>>>>>>>>>>>>>>>>>>>>>>>>>>>>>>>>>>>>

\section{Related Work}

\subsection{Molecular Design Models}

Molecular design models connect natural language and molecular structures through retrieval, translation, editing, or optimization.
Early work such as Text2Mol~\cite{edwards2021text2mol} learns joint text-molecule representations for caption-based retrieval, while later methods extend this line to bidirectional translation and in-context generation with LLMs~\cite{li2024empowering,li2024molreflect,liu2024conversational,liu2025rl}.
In parallel, chemical foundation models such as MolT5~\cite{edwards2022translation}, ChemDFM~\cite{zhao2025developing}, ChemDFM-R~\cite{zhao2025chemdfm}, and Chem-R~\cite{wang2025chem} improve text-level chemical competence through large-scale pretraining, reasoning supervision, distillation, and reinforcement learning.
Other methods further strengthen the text-structure interface through explicit alignment or structural guidance, such as MolXPT~\cite{liu-etal-2023-molxpt}, GeLLM$^3$O~\cite{dey-etal-2025-mathtt}, and MSR~\cite{jang-etal-2025-structural}.
Recent work has also begun exploring agentic systems for molecular design.
MT-Mol~\cite{kim2025mt} uses multi-agent, tool-guided reasoning for molecular optimization, MultiMol~\cite{yu2025collaborative} combines expert agents for multi-objective optimization, and PharmAgents~\cite{gao2025pharmagents} builds a broader multi-agent ecosystem for virtual pharmaceutical workflows.
While these methods show the promise of agent-based molecular discovery, they mainly focus on optimization, editing, or broader drug-discovery pipelines.
Our work is orthogonal to these model-centric advances: we treat such models as pluggable generators and focus on a debate-based framework that aims to orchestrate multiple models to better align text-structure intent with structural feasibility in molecular design.

\subsection{Multi-Agent Debate}

Multi-Agent Debate (MAD) enhances robustness and reasoning diversity by allowing multiple agents to generate, critique, and revise candidate outputs through adversarial interaction~\cite{du2023improving}.
BDoG~\cite{zheng2024picture} extends debate to multimodal settings by introducing blueprint-guided structures that organize evidence and reduce opinion collapse.
DRAG~\cite{hu2025removal} integrates debate into both retrieval and generation stages of RAG to mitigate compounding errors, reducing hallucination through role-specialized proponents, opponents, and judges.
Unlike prior MAD settings, our goal is to narrow the molecular text-structure gap utilizing multi-perspective reasoning and local structural evidence.

% %%%%%%%%%%%%%%%%%%%%%%%%%%%%%%%%%%%%%%%%%%%%%%%%%%%%%%%%%%%%%%%%%%%%%%%%%%%%%%%%%%%%%
% %%%%%%%%%%%%%%%%%%%%%%%%%%%%%%%%%%%%%%%%%%%%%%%%%%%%%%%%%%%%%%%%%%%%%%%%%%%%%%%%%%%%%
% %%%%%%%%%%%%%%%%%%%%%%%%%%%%%%%%%%%%%%%%%%%%%%%%%%%%%%%%%%%%%%%%%%%%%%%%%%%%%%%%%%%%%
%
% Method
%
% >>>>>>>>>>>>>>>>>>>>>>>>>>>>>>>>>>>>>>>>>>>>>>>>>>>>>>>>>>>>>>>>>>>>>>>>>>>>>>>>>>>>>

\section{Mol-Debate}
\label{sec:method}

In this section, we present \textbf{Mol-Debate}, an iterative generation framework designed to bridge the text-structure gap in molecular design,
with the overview shown in Figure~\ref{fig:pipeline}.
Due to the limited space, detailed implementation of Mol-Debate can be found in Appendix~\ref{sec:appendix}.

\subsection{Overview}
\label{sec:overview}

Formally, given a design instruction $\mathcal{X}$,
Mol-Debate maps $\mathcal{X}$ to a molecule $\mathcal{S}$, represented as a SMILES sequence~\cite{weininger1988smiles}:
\begin{align}
f:\,\,&\mathcal{X}\mapsto\,\mathcal{S}\nonumber\\
&f = f_{\text{Dev}} \circ f_{Exam} \circ f_{\text{Deb}} \circ f_{\text{Refiner}} \nonumber \\
    &\mathcal{X}=\{x_i\},\,\mathcal{S}=\{s_j\}.
\end{align}

Mol-Debate solves this mapping via an iterative multi-agent process $f$.
At round $r$, the state is
\begin{align}
\mathcal{T}^{r} = (\mathcal{X}^{r},\, \mathcal{P}^{r},\, \mathcal{A}),
\end{align}

where $\mathcal{X}^{r}$ is the current instruction, $\mathcal{P}^{r}$ is a pool of candidate molecules, and $\mathcal{A}$ denotes the agent set.
At the end of each round, both the pool and the instruction may be updated to produce $(\mathcal{X}^{r+1}, \mathcal{P}^{r+1})$, enabling progressive refinement.

\subsection{Developer-Debater Conflict}
\label{sec:dev_deb}

A key challenge is balancing chemistry-grounded proposal generation with flexible semantic adjudication.
Expert generators can produce chemically plausible candidates yet remain less adaptive to nuanced instruction language during critique.
Mol-Debate addresses this by decoupling agents: \textbf{Developer Agent} focuses on professional chemical exploration, while dedicated \textbf{Debate Agent} injects general reasoning to evaluate intent satisfaction and select instruction-aligned candidates.

\subsubsection{Developer Agent}
\label{sec:dev_agent}

To ensure candidates are grounded in professional chemistry principles, Mol-Debate uses a set of $N$ Developer agents.
Each Developer proposes a local subset of $M_n$ candidates $\mathcal{S}$ and provides an explicit design rationale $\mathcal{R}$.
\begin{align}
f_{\text{Dev}}:\,\, &\mathcal{X}^r \mapsto \mathcal{P}^{r} \nonumber\\
&\mathcal{P}^{r} = \bigcup_{n=1}^N \mathcal{P}_n
= \bigcup_{n=1}^N \bigcup_{m=1}^{M_n} \left\{ (\mathcal{S}_{n,m}, \mathcal{R}_{n,m}) \right\}.
\label{eq:dev}
\end{align}

Here, $\mathcal{P}_n$ denotes the molecule-rationale subset from the $n$-th Developer.
At this stage, the pool is designed for breadth, prioritizing chemically plausible exploration but not yet guaranteeing full constraint satisfaction.
Appendix~\ref{sec:app_prompts_for_agents} shows the prompt for LLMs in Developer Agent.

\subsubsection{Debate Agent}
\label{sec:debate_agent}

Given candidate pool $\mathcal{P}^r$, the Debate Agent coordinates $K$ Debaters and a Judge to select candidates $\mathcal{P}^{r+1}$ that best match the semantic intent of $\mathcal{X}^r$:
\begin{align}
f_{\text{Deb}}:\,\,&\mathcal{X}^r \times \mathcal{P}^r \mapsto \mathcal{P}^{r+1} \nonumber\\
&f_{\text{Deb}} = f_{\text{Debater}} \circ f_{\text{Judge}}.
\label{eq:debate_agent}
\end{align}

\noindent \textbf{Debater.}
The debating employs $K$ parallel Debater agents to critically evaluate the candidates against the structural evidence. Each debater analyzes the instruction $\mathcal{X}^r$, the candidate pool $\mathcal{P}^r$, and  selects a subset of sufficient molecules $\mathcal{P}^r_k$:
\begin{align}
f_{\text{Debater}}^{(k)}:\,\,&\mathcal{X}^r \times \mathcal{P}^r \mapsto \mathcal{P}_k^r \nonumber\\
&\mathcal{P}_k^r = \{ \mathcal{S} \mid (\mathcal{S},\mathcal{R}) \in \mathcal{P}^r \}.
\label{eq:debater_agent}
\end{align}

Appendix~\ref{sec:app_prompts_for_agents} shows the prompt for LLMs in Debater Agent.

\noindent \textbf{Judge.}
The Judge aggregates the debate results $\mathcal{D}^r=\{\mathcal{P}_k^r\}_{k=1}^K$ proposed by the $K$ independent Debaters.
The updated pool $\mathcal{P}^{r+1}$ is obtained by intersection when possible, with a union fallback:
\begin{align}
f_{\text{Judge}}:\,\,&\mathcal{D}^r \mapsto\,\mathcal{P}^{r+1}\nonumber\\
&\mathcal{P}^{r+1} = \begin{cases} 
\bigcap_{k=1}^K \mathcal{P}^r_k & \text{if } \bigcap \neq \emptyset \\
\bigcup_{k=1}^K \mathcal{P}^r_k & \text{otherwise}.
\end{cases}
\label{eq:judge}
\end{align}

This coupling ensures that candidate generation and intent matching are jointly enforced, thereby reducing the likelihood that the system accepts chemically plausible yet instruction-misaligned molecules.

\subsection{Global-Local Structural Reasoning}
\label{sec:glo_loc}

The second perspective bridges global textual representations and local chemical constraints.
While the Developer and Debate Agent can enforce intent, text-level debate alone cannot reliably detect local invalidity.
Mol-Debate therefore introduces an \textbf{Examiner Agent} that provides deterministic local structural evidence, and upgrades the Debate Agent to incorporate this evidence during selection.

\subsubsection{Examiner Agent}
\label{sec:exam_agent}

The Examiner Agent acts as an objective auditor using deterministic tools (e.g., RDKit~\cite{landrum2013rdkit}) to validate candidates and compute molecular properties.
It maps the candidate pool $\mathcal{P}^r$ to examination reports $\mathcal{E}$:
\begin{align}
f_{Exam}:\,\,&\mathcal{P}^r \mapsto \mathcal{E}^r \nonumber \\
&\mathcal{E}^r = \{e(\mathcal{S}) \mid (\mathcal{S}, \mathcal{R}) \in \mathcal{P}^r\} \nonumber\\
&e(\mathcal{S}) = \{(d, \phi_d(\mathcal{S})) \mid d \in \Pi \}.
\label{eq:exam}
\end{align}

The examination function $e()$ performs a check by property analysis $\phi()$. Let $\Pi$ be the set of chemical descriptors $d$.
This evidence anchors decision-making in verifiable chemical constraints,
providing a Local Structural Perspective for downstream Agents.
More implementation details of the chemical descriptors are provided in Appendix~\ref{sec:app_examiner_agent}.

\subsubsection{Local Structure-Grounded Debating}
\label{sec:debate_agent_upgraded}

With structural evidence $\mathcal{E}^r$ available, Mol-Debate upgrades debate to jointly consider semantic intent and local structural feasibility.

The $f_{\text{Deb}}$ in Eq.~(\ref{eq:debate_agent}) is upgraded as:
\begin{align}
f_{\text{Deb}}:\,\,&\mathcal{X}^r \times \mathcal{P}^r \times \textcolor{blue}{\mathcal{E}^r} \mapsto \mathcal{P}^{r+1},
\label{eq:debate_agent_full}
\end{align}

in which the $f_{\text{Debater}}$ in Eq.~(\ref{eq:debater_agent}) is upgraded as:
\begin{align}
f_{\text{Debater}}^{(k)}:\,\,&\mathcal{X}^r \times \mathcal{P}^r \times \textcolor{blue}{\mathcal{E}^r}\mapsto \mathcal{P}_k^r ,
\label{eq:debater_agent_full}
\end{align}

where each debater analyzes the instruction $\mathcal{X}^r$, the candidate pool $\mathcal{P}^r$ and structural examination report $\mathcal{E}^r$, then selects a subset of sufficient molecules $\mathcal{P}^r_k$.

\subsection{Static-Dynamic Integration}
\label{sec:sta_dyn}

The third perspective addresses the limitation of static generation.
Even with local structure-grounded debate, disagreement can persist due to underspecified instructions.
Mol-Debate therefore introduces a \textbf{Refiner Agent} that turns debate dissonance into a learning signal by reformulating the instruction.
The Refiner Agent analyzes the debate outcomes and structural evidence to diagnose ambiguity or missing constraints and produces an updated instruction:
\begin{align}
f_{\text{Refiner}}:\,\,&\mathcal{X}^r \times \mathcal{P}^r \times \mathcal{P}^{r+1} \times \mathcal{E}^r \times \mathcal{D}^r \mapsto \mathcal{X}^{r+1}.
\label{eq:refiner}
\end{align}
The refined instruction is fed back to the Developer Agent $f_{\text{Dev}}$ to generate improved candidates in the next round, closing a feedback loop.
Appendix~\ref{sec:app_prompts_for_agents} shows the prompt for LLMs in Refiner Agent.

% \subsection{Consensus Evaluation and Stopping Criterion}
\subsection{Consensus Evaluation and Termination}
\label{sec:consensus}

Mol-Debate iterates until either the debaters reach a sufficient level of agreement and the candidate pool collapses to a single candidate, or the maximum number of rounds $r_{\text{max}}$ is reached.
We quantify agreement using a consensus score $\mathcal{C}_r$ defined as the average similarity across subsets $\{\mathcal{P}_k^r\}_{k=1}^K$ selected by $K$ Debaters:
\begin{equation}
\mathcal{C}_r = \frac{1}{\binom{K}{2}} \sum_{i=1}^{K-1} \sum_{j=i+1}^{K} \frac{|\mathcal{P}^r_i \cap \mathcal{P}^r_j|}{|\mathcal{P}^r_i \cup \mathcal{P}^r_j|},
\label{eq:consensus_score}
\end{equation}

and define the stopping indicator $\mathcal{SI}$ as
\begin{equation}
f_{SI}:\mathcal{C}_r \times \mathcal{P}^{r+1} \mapsto \mathcal{SI}^r\nonumber
\end{equation}
\begin{equation}
\mathcal{SI}^r = 
\begin{cases} 
\text{1} & (\mathcal{C}_r \ge \theta \land |\mathcal{P}^{r+1}| = 1) \lor r \ge r_{\text{max}} \\
\text{0} & \text{otherwise}
\end{cases}.
\label{eq:stop}
\end{equation}

\subsection{Perspective-Oriented Orchestration}
\label{sec:orch}

Algorithm~\ref{alg:moldebate} summarizes the full workflow.
Across rounds, Mol-Debate first expands candidates via Developer Agents and prunes them via Debate Agent (Developer-Debater Conflict),
then grounds selection with deterministic local structural evidence from the Examiner Agent (Global-Local Structural Reasoning),
and finally resolves residual ambiguity through instruction reformulation by the Refiner Agent (Static-Dynamic Integration).
This orchestration enables reliable text-guided molecular design by iteratively coupling semantic intent, structural feasibility, and iterative refinement within a debate-driven agentic reasoning loop.
After termination of the debate, we output a single molecule via $\text{SelectOne()}$ by uniformly sampling one candidate from the final stabilized pool $\mathcal{P}^{r+1}$.

\begin{algorithm}
\caption{Mol-Debate Pipeline}
\label{alg:moldebate}
\begin{algorithmic}[1]
\REQUIRE $\mathcal{X}^0$, $K$, $r_\text{max}$, $\theta$
\ENSURE Optimal Molecule $\mathcal{S}^*$

\STATE $r \leftarrow 0$, $\mathcal{SI}^0 \leftarrow 0$, $\mathcal{P}^0 \leftarrow f_{\text{Dev}}(\mathcal{X}^0)$ \textcolor{gray}{// Eq.~(\ref{eq:dev})}

\WHILE{$\mathcal{SI}^r \neq 1$} \textcolor{gray}{// Eq.~(\ref{eq:stop})}

    \STATE $\mathcal{E}^r \leftarrow f_{\text{Exam}}(\mathcal{P}^r)$ \textcolor{gray}{// Eq.~(\ref{eq:exam})}
    
    \STATE $\mathcal{D}^r \leftarrow \{ f_{\text{Debater}}^{(k)}(\mathcal{X}^r, \mathcal{P}^r, \mathcal{E}^r) \}_{k=1}^K$ \textcolor{gray}{//Eq.~(\ref{eq:debater_agent_full})}

    \STATE $\mathcal{P}^{r+1} \leftarrow f_{\text{Judge}}(\mathcal{D}^r)$ \textcolor{gray}{// Eq.~(\ref{eq:judge})}

    \IF{$\mathcal{C}_r < \theta$} \textcolor{gray}{// Eq.~(\ref{eq:consensus_score})}
        \STATE \textcolor{gray}{// Eq.~(\ref{eq:refiner})}
        \STATE $\mathcal{X}^{r+1} \leftarrow f_{\text{Refiner}}(\mathcal{X}^r, \mathcal{P}^r, \mathcal{P}^{r+1},  \mathcal{E}^r, \mathcal{D}^r)$
        \STATE $\mathcal{P}^{r+1} \leftarrow \mathcal{P}^{r+1} \cup f_{\text{Dev}}(\mathcal{X}^{r+1})$ \textcolor{gray}{// Eq.~(\ref{eq:dev})}
    \ELSE
        \STATE $\mathcal{X}^{r+1} \leftarrow \mathcal{X}^{r}$
    \ENDIF
    \STATE $r \leftarrow r + 1$
\ENDWHILE

\STATE $\mathcal{S}^* \leftarrow \text{SelectOne} (\mathcal{P}^{r+1})$
\end{algorithmic}
\end{algorithm}

% %%%%%%%%%%%%%%%%%%%%%%%%%%%%%%%%%%%%%%%%%%%%%%%%%%%%%%%%%%%%%%%%%%%%%%%%%%%%%%%%%%%%%
% %%%%%%%%%%%%%%%%%%%%%%%%%%%%%%%%%%%%%%%%%%%%%%%%%%%%%%%%%%%%%%%%%%%%%%%%%%%%%%%%%%%%%
% %%%%%%%%%%%%%%%%%%%%%%%%%%%%%%%%%%%%%%%%%%%%%%%%%%%%%%%%%%%%%%%%%%%%%%%%%%%%%%%%%%%%%
%
% Experimental Setup
%
% >>>>>>>>>>>>>>>>>>>>>>>>>>>>>>>>>>>>>>>>>>>>>>>>>>>>>>>>>>>>>>>>>>>>>>>>>>>>>>>>>>>>>

\begin{table*}[t]
    \centering
    \adjustbox{max width=\linewidth}{
    \begin{tabular}{llcccccccc}
        \toprule
         Method & Approach & EM $\uparrow$ & BLEU $\uparrow$ &		Lev. $\downarrow$ &	MA.F $\uparrow$ &	RD.F $\uparrow$ &	MO.F $\uparrow$ &	FCD $\downarrow$ &	Val. $\uparrow$ \\
        \midrule
         Llama-3.1-8B-Instruct (Meta, 2024) & Naive Gen & 0.0045 & 0.1873 &		73.8242 &	0.5838 &	0.3489 &	0.2644 &	6.9835 &	0.4167 \\
         GPT-4o mini (OpenAI, 2024) & Naive Gen & 0.0173 & 0.1947 &		63.2667 &	0.5988 &	0.3629 &	0.3027 &	8.6415 &	0.5148 \\
         GPT-5 mini (OpenAI, 2025) & Naive Gen & 0.1773 & 0.4046 &	47.1945 &	0.8829 &	0.7280 &	0.6575 &	1.2423 &	0.7942 \\
         Gemma-4-31b-it (Google, 2026) & Naive Gen &  0.0452 & 0.3786	&	49.8358 &	0.7067 &	0.4932 &	0.4169 &	3.8080 &	0.7061 \\ 
        \midrule
        MolT5-large (EMNLP, 2022) & PT+FT & 0.3110 &	0.8540 & 16.0710 &	0.8340 &	0.7460 &	0.6840 &	1.2000 &	0.9050 \\
        BioT5-base (EMNLP, 2023) & PT+FT & 0.4130 &	\colorbox{MyLightBlue}{0.8670} &	\colorbox{MyLightBlue}{15.0970} &	0.8860 &	0.8010 &	0.7340	 & 0.4300	 & \colorbox{MyBlue}{\textbf{1.0000}} \\
        MolReGPT (TKDE, 2024) & RAG & 0.2800 & 0.8570 &	17.1400 &	0.9030 &	0.8050 &	0.7390 &	0.5900 &	0.8990 \\
        Mol-Instruction (ICLR, 2024) & FT & 0.0400 & 0.3000 &	39.4200 &	0.4400 &	0.2900 &	0.2500  & - & \colorbox{MyBlue}{\textbf{1.0000}} \\
        Mol-R1 (arXiv, 2025) & CoT+Distill+RL & 0.2340 &	0.6410 &	32.9400 &	0.8230 &	0.6840 &	0.6120 & - & 0.8470 \\
        MSR (ACL, 2025) & CoT+FT & 0.4210 & \colorbox{MyBlue}{\textbf{0.8780}} &	\colorbox{MyBlue}{\textbf{12.7600}} &	0.9240 &	0.8560 &	0.8040 &	0.2600 & 0.9820 \\
         ChemDFM-v1.5-8B (Cell Press, 2025) & PT+FT & \colorbox{MyLightBlue}{0.5379} & 0.6086 & 41.1333 &	\colorbox{MyLightBlue}{0.9388} &	\colorbox{MyLightBlue}{0.8798} &	\colorbox{MyLightBlue}{0.8314} &	\colorbox{MyLightBlue}{0.1957} &	0.9770 \\
          ChemDFM-R-14B (arXiv, 2025)  & PT+CoT+FT & 0.4948 & 0.6733 &		33.9506 &	0.9313 &	0.8592 &	0.8092 &	0.2236 &	0.9733 \\
          Chem-R-8B (arXiv, 2025)  & CoT+FT+RL & 0.4385 & 0.8648 & 16.1830 &	0.9204 &	0.8398 &	0.7827 &	0.2747 &	0.9664 \\

        \midrule
        \textbf{Mol-Debate (Ours)} & Mol-Debate & \colorbox{MyBlue}{\textbf{0.5982}} &	0.7059 &	30.8294 &	\colorbox{MyBlue}{\textbf{0.9601}} &	\colorbox{MyBlue}{\textbf{0.9103}} &	\colorbox{MyBlue}{\textbf{0.8694}} &	\colorbox{MyBlue}{\textbf{0.1258}} &	\colorbox{MyLightBlue}{0.9973} \\
         \bottomrule
    \end{tabular}
    }
    \caption{Overall performance of Mol-Debate and other baselines on caption-to-molecule generation task. \textcolorblue{\textbf{Blue}} marks the \textcolorblue{\textbf{best-performing}} method, \textcolorlblue{light blue} represents the \textcolorlblue{second-best-performing} method.}
    \label{tab:c2m}
\end{table*}

\begin{table*}[t]
    \centering
    % \footnotesize
    \adjustbox{max width=\linewidth}{
    \begin{tabular}{llcc|cc|cc|cc}
        \toprule
        \multirow{2}{*}{Method}& \multirow{2}{*}{Approach} & \multicolumn{2}{c}{MolCustom} & \multicolumn{2}{c}{MolEdit} & \multicolumn{2}{c}{MolOpt}  & \multicolumn{2}{c}{\textbf{Average}}\\
        \cmidrule(lr){3-10}
          & & SR $\uparrow$ & WSR $\uparrow$ & SR $\uparrow$ & WSR $\uparrow$ & SR $\uparrow$ & WSR $\uparrow$ & SR $\uparrow$ & WSR $\uparrow$\\
        \midrule
            Llama-3.1-8B-Instruct (Meta, 2024) & Naive Gen & 0.0471  & 0.0291 &  0.2951 & 0.1605 & 0.3394  & 0.1341 &  0.2272 & 0.1079  \\
            GPT-4o mini (OpenAI, 2024) & Naive Gen &  0.1063 & 0.0696 &  0.5059 & 0.3549 & 0.3574  & 0.2727 &   0.3232 & 0.2324 \\
            GPT-5 mini (OpenAI, 2025) & Naive Gen &  \colorbox{MyBlue}{\textbf{0.5045}} & 		\colorbox{MyBlue}{\textbf{0.3323}} & \colorbox{MyLightBlue}{0.8865} &	\colorbox{MyLightBlue}{0.6033} & 0.8454 &	0.4069 & \colorbox{MyLightBlue}{0.7455} &	\colorbox{MyLightBlue}{0.4475} \\
            Gemma-4-31b-it (Google, 2026) & Naive Gen & 0.1876 &	0.1217 & 0.7291 &	0.5003 & 0.6933 &	0.5156 & 0.5366 &	0.3792 \\
        \midrule
        MolT5-large (EMNLP, 2022) & PT+FT &  0.0217 &	0.0135  &   0.2251  & 0.0244  &   0.4465 & 0.0489 & 0.2311 & 0.0289\\
        BioT5-base (EMNLP, 2023) & PT+FT &  0.0224 &	0.0158 &  0.1938 &	0.0306  & 0.5095  &	0.0799 & 0.2419	& 0.0421 \\
            ChemDFM-v1.5-8B (Cell Press, 2025) & PT+FT &  0.0835 & 0.0560 &  0.3401 & 0.2571 & 0.3876 & 0.2934 &   0.2704 &	0.2022  \\
            ChemDFM-R-14B (arXiv, 2025) & PT+CoT+FT & 0.1001 & 0.0662 &  0.4845 & 0.3317 &  0.2658 & 0.1849 &   0.2835 &	0.1943  \\
            Chem-R-8B (arXiv, 2025) & CoT+FT+RL & 0.2107  & 0.1419 & 0.7955  & 0.5429 &  \colorbox{MyLightBlue}{0.8561} & \colorbox{MyLightBlue}{0.5789} &   0.6208 & 0.4213 \\

        \midrule
          \textbf{Mol-Debate (Ours)} & Mol-Debate &  \colorbox{MyLightBlue}{0.3965} &	\colorbox{MyLightBlue}{0.2671} & \colorbox{MyBlue}{\textbf{0.9399}} &	\colorbox{MyBlue}{\textbf{0.6432}} & \colorbox{MyBlue}{\textbf{0.9203}} &	\colorbox{MyBlue}{\textbf{0.6052}} & \colorbox{MyBlue}{\textbf{0.7522}} & \colorbox{MyBlue}{\textbf{0.5052}} \\

         \bottomrule
    \end{tabular}
    }
    \caption{Overall performance of Mol-Debate and other baselines on the text-based open molecule generation task. 
    }
    \label{tab:tomg}
\end{table*}

\section{Experimental Setup}

\subsection{Datasets}

% \noindent \textbf{ChEBI-20} 

We evaluate caption-to-molecule generation on \textbf{ChEBI-20}~\cite{edwards2021text2mol}, which contains 33010 molecule-caption pairs and requires generating a SMILES string from a textual description.
We follow the evaluation protocol in~\cite{edwards2022translation}, reporting SMILES exact matches (EM), BLEU scores, Levenshtein distance (Lev.)~\cite{miller2009levenshtein}, fingerprint similarity (MA.F, RD.F, MO.F), Fréchet ChemNet Distance (FCD), and SMILES validity (Val.).
We further evaluate open-ended text-guided molecule generation on \textbf{S$^2$-Bench}~\cite{li2025speaktostructureevaluatingllmsopendomain}, which includes 5000 queries per subtask for customized molecule generation (MolCustom), editing (MolEdit), and optimization (MolOpt), resulting in 15000 queries totally.
We report the Success Rate (SR) and Weighted Success Rate (WSR).

\subsection{Baselines}

We evaluate the performance of our method with various LLMs.
We employ general LLMs, Llama-3.1-8B-Instruct~\cite{dubey2024llama}, GPT-4o mini~\cite{hurst2024gpt}, GPT-5 mini~\cite{singh2025openai} and Gemma-4-31b-it~\cite{gemma4}.
We use chemical LLMs, 
MolT5-large~\cite{edwards2022translation},
BioT5-base~\cite{pei2023biot5},
MolReGPT~\cite{li2024empowering},
Mol-Instruction~\cite{fangmol},
Mol-R1~\cite{li2025mol},
MSR~\cite{jang-etal-2025-structural},
ChemDFM-v1.5-8B~\cite{zhao2025developing},
ChemDFM-R-14B~\cite{zhao2025chemdfm}, 
and Chem-R-8B~\cite{wang2025chem}.

\subsection{Implementation Details}
The maximum debate round $r_\text{max}$ is set to 4. The number of debaters $K$ is set to 2. The consensus score threshold $\theta$ is set to 0.6.
Mol-Debate uses ChemDFM-R-14B, ChemDFM-v1.5-8B, and Chem-R-8B as LLMs in the Developer Agent. 
For the main results in Tables~\ref{tab:c2m} and~\ref{tab:tomg}, Mol-Debate uses GPT-5 mini as the LLM for the Debate Agent and Refiner Agent.
For the ablations in Tables~\ref{tab:e-g_components} and~\ref{tab:ablation}, and Figure~\ref{fig:round_eval}, Mol-Debate uses GPT-4o mini for the Debate Agent and Refiner Agent.
More implementation details can be found in Appendix~\ref{sec:app_implementation_details}.

% %%%%%%%%%%%%%%%%%%%%%%%%%%%%%%%%%%%%%%%%%%%%%%%%%%%%%%%%%%%%%%%%%%%%%%%%%%%%%%%%%%%%%
% %%%%%%%%%%%%%%%%%%%%%%%%%%%%%%%%%%%%%%%%%%%%%%%%%%%%%%%%%%%%%%%%%%%%%%%%%%%%%%%%%%%%%
% %%%%%%%%%%%%%%%%%%%%%%%%%%%%%%%%%%%%%%%%%%%%%%%%%%%%%%%%%%%%%%%%%%%%%%%%%%%%%%%%%%%%%
%
% Experimental Results
%
% >>>>>>>>>>>>>>>>>>>>>>>>>>>>>>>>>>>>>>>>>>>>>>>>>>>>>>>>>>>>>>>>>>>>>>>>>>>>>>>>>>>>>

\begin{figure*}[t]
    \centering
    \includegraphics[width=0.9\linewidth]{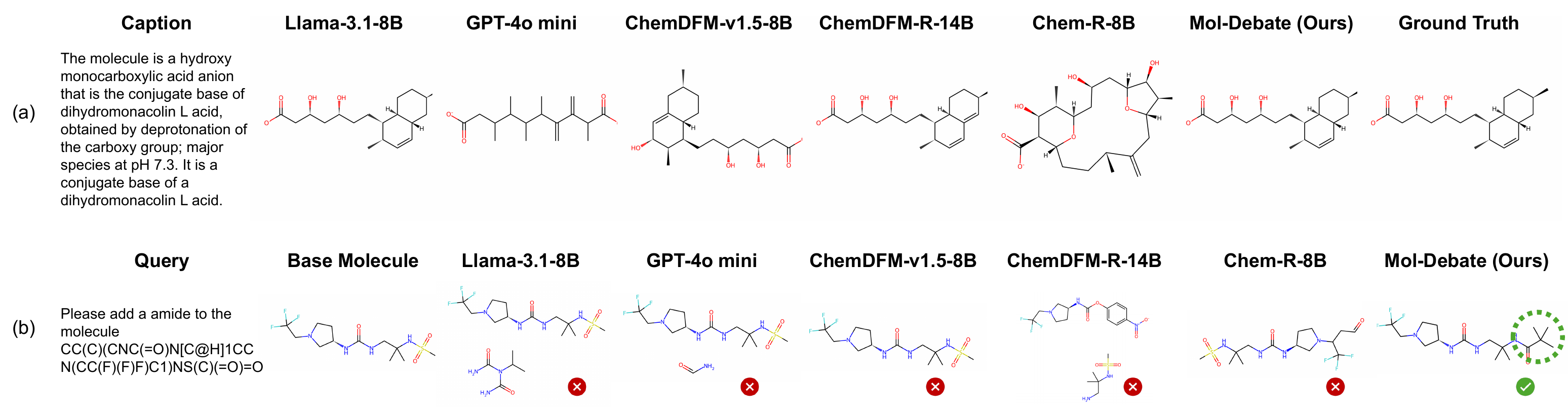}
    \caption{Examples of samples from (a) caption-to-molecule and (b) text-based open molecule generation task.}
    \label{fig:compare_two}
\end{figure*}

\section{Experimental Results}

\subsection{Comparison with Baselines}

\noindent \textbf{Caption-to-molecule generation task on ChEBI-20.}
As shown in Table~\ref{tab:c2m}, Mol-Debate achieves the best EM of 0.5982, together with the strongest overall similarity scores and a competitive validity score.
Compared with the second-best method, Mol-Debate improves EM by an absolute 6.03 percentage points.
Figure~\ref{fig:compare_two}(a) presents example molecules generated by Mol-Debate.

\noindent \textbf{Open-domain molecule generation task on S$^2$-Bench.}
Table~\ref{tab:tomg} evaluates more realistic discovery-oriented tasks.
Mol-Debate achieves the best average performance, reaching 0.7522 in SR and 0.5052 in WSR.
Compared with the second-best method, Mol-Debate improves WSR by an absolute 5.77 percentage points.
Examples in Figure~\ref{fig:compare_two}(b) further illustrate the effectiveness of Mol-Debate.

More generated samples can be found in Appendix~\ref{sec:app_samples}.

\begin{table}[h]
    \centering
    % \footnotesize
    \adjustbox{max width=\linewidth}{
        \begin{tabular}{lccccc}
            \toprule
             Settings & EM $\uparrow$ & BLEU $\uparrow$ & MO.F $\uparrow$ &	FCD $\downarrow$ &	Val. $\uparrow$ \\
            \midrule
              \textbf{D+G (Ours)}  &	\colorbox{MyBlue}{\textbf{0.5542}} & \colorbox{MyBlue}{\textbf{0.7008}}  &	\colorbox{MyBlue}{\textbf{0.8362}} &	\colorbox{MyBlue}{\textbf{0.1850}} &	\colorbox{MyBlue}{\textbf{1.0000}} \\
              D+D  & 	 \colorbox{MyLightBlue}{0.4942} &	\colorbox{MyLightBlue}{0.6996} &	\colorbox{MyLightBlue}{0.8057} &	\colorbox{MyLightBlue}{0.2057} &	\colorbox{MyLightBlue}{0.9988} \\
             G+G  &	0.0206 & 0.1188 & 0.2825 &	6.1839 &	0.8791 \\
             \midrule
               \textbf{Mol-Debate (Ours)} &	\colorbox{MyBlue}{\textbf{0.5542}} & 0.7008 & \colorbox{MyBlue}{\textbf{0.8362}} &	\colorbox{MyBlue}{\textbf{0.1850}} &	\colorbox{MyBlue}{\textbf{1.0000}} \\
                    
                $-$ E &	0.5030 & 	\colorbox{MyBlue}{\textbf{0.7238}} & 0.8227 &	0.2139 &	0.9748 \\
                    
                $-$ R & \colorbox{MyLightBlue}{0.5324} & \colorbox{MyLightBlue}{0.7221} &	\colorbox{MyLightBlue}{0.8279} &	\colorbox{MyLightBlue}{0.1894} &	\colorbox{MyBlue}{\textbf{1.0000}} \\
                    
                $-$ E, $-$ R &	0.5009 & 0.7190 & 0.8185 &	0.2146 &	0.9730 \\
             \bottomrule
        \end{tabular}
    }
    \caption{Ablation study on the developer-generalist LLMs debating (upper) and key components (lower) of Mol-Debate on the caption-to-molecule generation task. D: Developer LLM, G: Generalist LLM. E: Examiner Agent, R: Refiner Agent.
    }
   
    \label{tab:e-g_components}
\end{table}

\begin{figure}[http]
    \centering
    \includegraphics[width=1\linewidth]{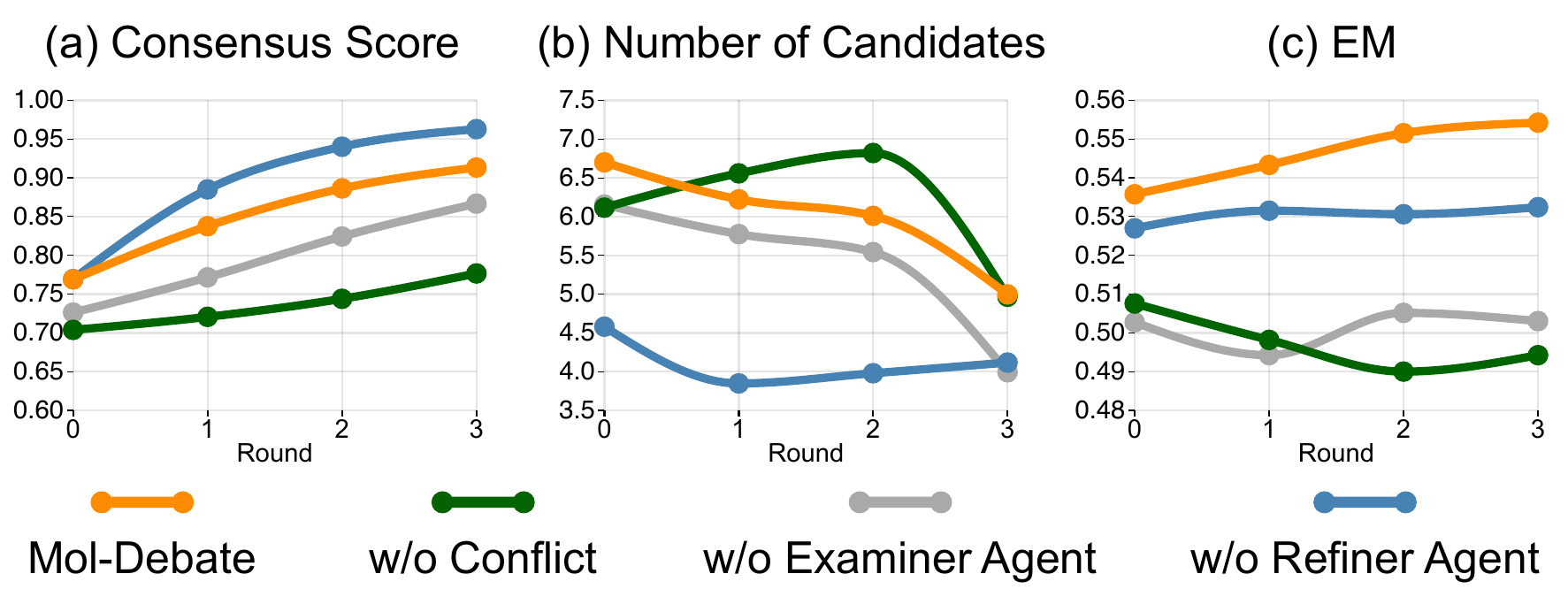}
    \caption{Analysis of (a) consensus score $\mathcal{C}_r$, (b) the number of candidates $|\mathcal{P}^{r}|$, and (c) EM score with respect to the round of debate.}
    \label{fig:round_eval}
    
\end{figure}

\subsection{Analysis of Multiple Perspectives}

\noindent \textbf{Developer-Debater skill conflict enhances structural reasoning.}
Table~\ref{tab:e-g_components} shows that the best results come from mixing LLMs with distinct skills rather than using homogeneous agents.
A generalist-only debate (G+G, solely uses GPT-4o mini as LLM) performs poorly, indicating that language-only critique is insufficient under chemical constraints.
A developer-only debate (D+D, replace GPT-4o mini with ChemDFM-R) yields lower EM, suggesting developers can converge on valid yet instruction-misaligned candidates. 
Figure~\ref{fig:round_eval} further supports that the developer-only setting (w/o Conflict) maintains the lowest consensus score across rounds and its EM degrades as debate proceeds.

\begin{figure*}[t]
    \centering
    \includegraphics[width=0.9\linewidth]{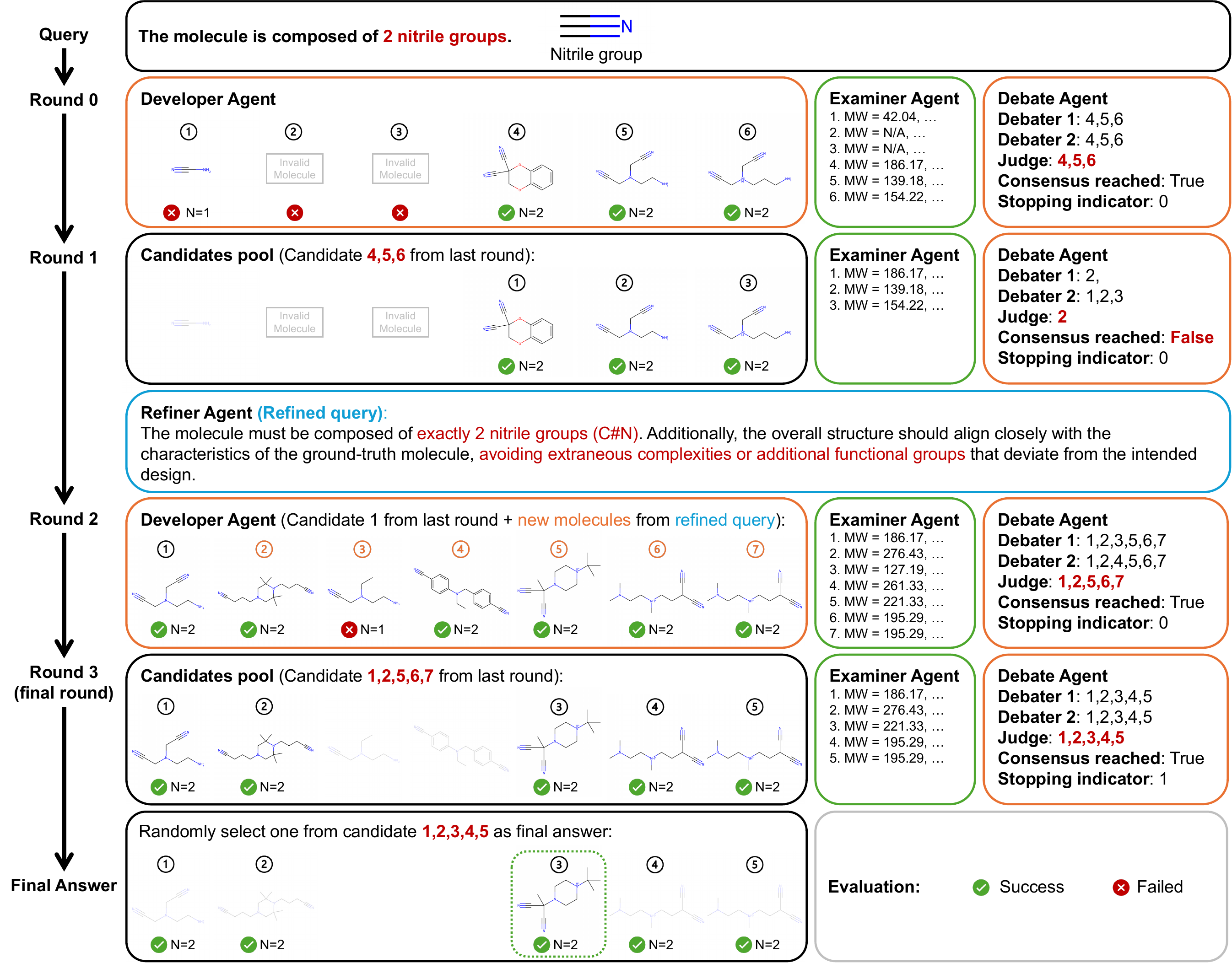}
    \caption{Case study of the Mol-Debate.
    In Round~0, the Developer Agent proposes 6 candidates and the Examiner Agent reports their examination results, providing local structural evidence for downstream debate.
    Using this evidence, the Debaters exclude candidates with low reliability (1, 2 and 3).
    Round~1 fails to reach consensus, triggering the Refiner to clarify underspecified constraints.
    Conditioned on the refined query, Round~2 expands the pool and the Debate Agent further removes intent-misaligned candidates (3).
    By the final round, Mol-Debate converges to a reliable candidate set and outputs a successful molecule.
    }
    \label{fig:tomg_case}
\end{figure*}

\noindent \textbf{Global-local structural evidence reduces chemical hallucinations.}
Table~\ref{tab:e-g_components} shows that removing the Examiner Agent leads to a clear drop in EM and validity.
Consistent with this, Figure~\ref{fig:round_eval} shows lower and slower-growing consensus scores $\mathcal{C}_r$ across rounds, indicating that debaters struggle to agree when local structural evidence is missing.
Overall, deterministic examination provides the Local Structural Perspective that anchors debate beyond global, text-level plausibility.

\noindent \textbf{Static-dynamic refinement improves semantic intent alignment through iterative exploration.}
Removing the Refiner Agent reduces EM while largely preserving validity, as shown in Table~\ref{tab:e-g_components}, implying that refinement primarily affects intent satisfaction rather than chemical correctness.
Figure~\ref{fig:round_eval} further supports this: without refinement, the candidate pool remains smaller throughout the rounds, limiting exploration on intent-consistent design options.
In contrast, Mol-Debate maintains a larger pool in early rounds and steadily increases both $\mathcal{C}_r$ and EM, illustrating how refinement guides the system toward semantically correct structures.

More analysis can be found in Appendix~\ref{sec:app_more_analysis}.
Case study in Figure~\ref{fig:tomg_case} shows how Mol-Debate cooperates with each agent from three perspectives.
More case studies can be found in Appendix~\ref{sec:app_case}.

\begin{table}[h]
    \centering
    % \footnotesize
    \adjustbox{max width=\linewidth}{
        \begin{tabular}{cccccc}
            \toprule
             Settings & EM $\uparrow$ & BLEU $\uparrow$ & MO.F $\uparrow$ &	FCD $\downarrow$ &	Val. $\uparrow$ \\
            \midrule
              $K$=2  &	\colorbox{MyBlue}{\textbf{0.5542}} & 0.7008  &	\colorbox{MyBlue}{\textbf{0.8362}} &	\colorbox{MyLightBlue}{0.1850} &	\colorbox{MyBlue}{\textbf{1.0000}} \\
              $K$=3  & 	\colorbox{MyLightBlue}{0.5485} & \colorbox{MyLightBlue}{0.7085} &	\colorbox{MyLightBlue}{0.8351}  &	0.1857 &	\colorbox{MyBlue}{\textbf{1.0000}} \\
             $K$=4  & 0.5373 & \colorbox{MyBlue}{\textbf{0.7157}} &	0.8333 &	\colorbox{MyBlue}{\textbf{0.1807}} &	0.9997 \\
             \midrule
             $R$=1 & 0.5358 & 0.7007 & 0.8274 & \colorbox{MyBlue}{\textbf{0.1843}} & \colorbox{MyBlue}{\textbf{1.0000}} \\
             
             $R$=2  &  0.5433 & \colorbox{MyBlue}{\textbf{0.7039}} &	0.8280 &	0.1878 &	\colorbox{MyBlue}{\textbf{1.0000}} \\
             
             $R$=3 &  \colorbox{MyLightBlue}{0.5515} & 0.6978 &	\colorbox{MyLightBlue}{0.8332} &	0.1894 &	\colorbox{MyBlue}{\textbf{1.0000}} \\
              
              $R$=4  &	\colorbox{MyBlue}{\textbf{0.5542}} & \colorbox{MyLightBlue}{0.7008} &	\colorbox{MyBlue}{\textbf{0.8362}} &	\colorbox{MyLightBlue}{0.1850} &	\colorbox{MyBlue}{\textbf{1.0000}} \\
             
             \midrule
              $\theta$=0.4  &	0.5361  & \colorbox{MyBlue}{\textbf{0.7063}} &	0.8285  &	\colorbox{MyLightBlue}{0.1837}  &	\colorbox{MyLightBlue}{0.9997} \\
              
              $\theta$=0.6  &	\colorbox{MyBlue}{\textbf{0.5542}} & 0.7008 & \colorbox{MyBlue}{\textbf{0.8362}} &	0.1850 &	\colorbox{MyBlue}{\textbf{1.0000}} \\
              
              $\theta$=0.8  &	\colorbox{MyLightBlue}{0.5485}  &	\colorbox{MyLightBlue}{0.7040}  &	\colorbox{MyLightBlue}{0.8331}  &	\colorbox{MyBlue}{\textbf{0.1820}}  &	\colorbox{MyLightBlue}{0.9997} \\
             \bottomrule
        \end{tabular}
    }
    \caption{Ablation study of Mol-Debate on the number of debaters ($K$), debate rounds ($R$) and consensus threshold ($\theta$) on the caption-to-molecule generation task.
    }
    \label{tab:ablation}
\end{table}

\subsection{Ablation Study}

We study the sensitivity of Mol-Debate to three hyperparameters as shown in Table~\ref{tab:ablation}.
It is found that a small number of debaters $K$ performs best, while larger $K$ yields only minor changes and slightly weaker exact recovery.
Increasing debate rounds $R$ steadily improves EM with validity remaining perfect, supporting the benefit of iterative refinement.
A moderate consensus threshold $\theta$ provides the best overall trade-off. While looser or stricter thresholds slightly degrade performance.
This finding suggests that effective molecular discovery requires a balance between rigorous agreement and the preservation of chemical diversity.
More ablation studies can be found in Appendix~\ref{sec:app_more_ablation_results}.

% \subsection{Case Study}

% \todo

% %%%%%%%%%%%%%%%%%%%%%%%%%%%%%%%%%%%%%%%%%%%%%%%%%%%%%%%%%%%%%%%%%%%%%%%%%%%%%%%%%%%%%
% %%%%%%%%%%%%%%%%%%%%%%%%%%%%%%%%%%%%%%%%%%%%%%%%%%%%%%%%%%%%%%%%%%%%%%%%%%%%%%%%%%%%%
% %%%%%%%%%%%%%%%%%%%%%%%%%%%%%%%%%%%%%%%%%%%%%%%%%%%%%%%%%%%%%%%%%%%%%%%%%%%%%%%%%%%%%
%
% Conclusion
%
% >>>>>>>>>>>>>>>>>>>>>>>>>>>>>>>>>>>>>>>>>>>>>>>>>>>>>>>>>>>>>>>>>>>>>>>>>>>>>>>>>>>>>

\section{Conclusion}

We study text-guided molecular design, where natural-language intent should be translated into a chemically valid structure under strict constraints, and identify the key challenge as a persistent text-structure gap.
We propose Mol-Debate, an iterative framework integrating perspectives in a generate-debate-refine loop: the Developer Agent explores chemically grounded candidates, the Examiner Agent provides local structural evidence, the Debate Agent adjudicates intent under evidence, and the Refiner Agent reformulates the instruction.
Experiments show that Mol-Debate achieves consistently high validity and outperforms strong general and chemical LLM baselines.

% %%%%%%%%%%%%%%%%%%%%%%%%%%%%%%%%%%%%%%%%%%%%%%%%%%%%%%%%%%%%%%%%%%%%%%%%%%%%%%%%%%%%%
% %%%%%%%%%%%%%%%%%%%%%%%%%%%%%%%%%%%%%%%%%%%%%%%%%%%%%%%%%%%%%%%%%%%%%%%%%%%%%%%%%%%%%
% %%%%%%%%%%%%%%%%%%%%%%%%%%%%%%%%%%%%%%%%%%%%%%%%%%%%%%%%%%%%%%%%%%%%%%%%%%%%%%%%%%%%%
%
% Limitations
%
% >>>>>>>>>>>>>>>>>>>>>>>>>>>>>>>>>>>>>>>>>>>>>>>>>>>>>>>>>>>>>>>>>>>>>>>>>>>>>>>>>>>>>

\section*{Limitations}

Mol-Debate performs multi-agent, multi-round generation, debate, and examination, leading to higher inference cost than single-pass baselines.
This overhead can be substantial, particularly in interactive settings.
Such efficiency trade-offs are common in multi-agent and reasoning-based systems, and remain an open question.
A promising direction for future work is to develop stronger Judge policies and tighter early-stopping criteria to avoid unnecessary rounds and accelerate inference.

The Examiner Agent currently provides deterministic validity checks and basic property profiling.
While this evidence is essential for preventing chemically invalid outputs, it does not guarantee higher-level objectives such as synthesizability and bioactivity.
Future work can extend the Examiner with stronger or task-specific evaluators and propagate this richer evidence into debate.

For underspecified instructions, multiple distinct molecules may legitimately satisfy the prompt.
In such cases, the final output can depend on debate preferences and the granularity of success criteria.
A key future direction is to make selection more controllable by exposing explicit preferences, thereby enabling downstream selection in practical workflows.

% %%%%%%%%%%%%%%%%%%%%%%%%%%%%%%%%%%%%%%%%%%%%%%%%%%%%%%%%%%%%%%%%%%%%%%%%%%%%%%%%%%%%%
% %%%%%%%%%%%%%%%%%%%%%%%%%%%%%%%%%%%%%%%%%%%%%%%%%%%%%%%%%%%%%%%%%%%%%%%%%%%%%%%%%%%%%
% %%%%%%%%%%%%%%%%%%%%%%%%%%%%%%%%%%%%%%%%%%%%%%%%%%%%%%%%%%%%%%%%%%%%%%%%%%%%%%%%%%%%%
%
% Acknowledgments
%
% >>>>>>>>>>>>>>>>>>>>>>>>>>>>>>>>>>>>>>>>>>>>>>>>>>>>>>>>>>>>>>>>>>>>>>>>>>>>>>>>>>>>>

\section*{Acknowledgments}

This work was supported by the National Natural Science Foundation of China (NSFC) under Grant No. T2541073 and No. 62372314.
This work was also supported by computational resources provided by The Centre for Large AI Models (CLAIM) of The Hong Kong Polytechnic University.
Wengyu Zhang expresses sincere gratitude to Dr. Changmeng Zheng, Yiyang Jiang, Wentao Hu, Jinhao Shen, and Yining Lu for their support.

\bibliography{custom}

\appendix

\newpage

\section{Appendix}
\label{sec:appendix}

\subsection{More Implementation Details}
\label{sec:app_implementation_details}

The implementation of our method and the baselines are built upon the Python code from the MolReGPT~\cite{li2024empowering} and S$^2$-Bench~\cite{li2025speaktostructureevaluatingllmsopendomain}.
We run the experiments on eight NVIDIA H20 GPU devices.
For proprietary LLMs, we use the OpenAI API framework\footnote{https://platform.openai.com/}. For open-source LLMs, we use the vLLM framework\footnote{https://docs.vllm.ai/}.

For the caption-to-molecule generation task on ChEBI-20, Mol-Debate uses ChemDFM-R-14B, ChemDFM-v1.5-8B, and Chem-R-8B as LLMs in the Developer Agent, and the number of generated candidates $M_n$ for each LLM is 4, 2, and 2, respectively. 
For the text-based open molecule generation task on S$^2$-Bench, Mol-Debate uses ChemDFM-R-14B and Chem-R-8B as LLMs in the Developer Agent, and the number of generated candidates $M_n$ for each LLM is 3.

For the results in Tables~\ref{tab:c2m} and~\ref{tab:tomg}, Mol-Debate uses GPT-5 mini as the LLM for the Debate Agent and Refiner Agent.
For the ablations in Tables~\ref{tab:e-g_components},~\ref{tab:ablation},~\ref{tab:app_developer_generalist_tomg},~\ref{tab:app_key_components_c2m},~\ref{tab:app_key_components_tomg},~\ref{tab:app_debater_number},~\ref{tab:app_debate_rounds},~\ref{tab:app_threshold} and Figure~\ref{fig:round_eval}, Mol-Debate uses GPT-4o mini for the Debate Agent and Refiner Agent.

For Table~\ref{tab:c2m}, results for MolT5-large, BioT5-base, MolReGPT, and Mol-R1 are sourced from their original papers, and Mol-Instruction results are sourced from~\cite{wang2025chem}.
For Table~\ref{tab:tomg}, results for MolT5-large and BioT5-base are sourced from~\cite{li2025speaktostructureevaluatingllmsopendomain}.
All remaining results in Tables~\ref{tab:c2m} and~\ref{tab:tomg} are reproduced by us under the corresponding evaluation protocols.

For the configuration of LLM inference, the temperature is set to 1, top p is set to 0.8, random seed is set to 42, max new tokens is set to 4096 for reasoning LLM, and 1024 for non-reasoning LLM.

\subsection{Prompts for Agents}
\label{sec:app_prompts_for_agents}

\begin{tcolorbox}[colback=black!1!white,colframe=black!57!white,title=Prompt for Developer Agent]

You are now working as an excellent expert in chemisrty and drug discovery. You always reason thoroughly before giving response. The reasoning process and answer are enclosed within <think> </think> and <answer> </answer> tags, respectively.i.e.,<think>reasoning process here</think><answer>answer here</answer>. Given the caption of a molecule, your job is to predict the SMILES representation of the molecule. The molecule caption is a sentence that describes the molecule, which mainly describes the molecule's structures, properties, and production. You can infer the molecule SMILES representation from the caption.\\
Task Format: \\
Instruction: Given the caption of a molecule, predict the SMILES representation of the molecule.\\
Input: [CAPTION\_MASK]\\
Your response should only be in the format above.\\
\{query\}
\end{tcolorbox}

\begin{tcolorbox}[colback=black!1!white,colframe=black!57!white,title=Prompt for Debater Agent, breakable]

You are an evaluator in caption to molecule task. Given the following candidate molecule SMILES of the input caption and the corresponding reason and examining results, evaluate the sufficiency of the molecules and select one or multiple sufficient molecules. Deliver a short, brief, and strong argument. The thinking process and argument are enclosed within <think> </think> and <answer> </answer> tags, respectively. i.e., <think> thinking process here </think> <answer> answer here </answer>.\\
Input caption: \{query\}\\
Candidate molecules: \{molecules\}\\
Output comma-separated index(s) of the sufficient molecule(s), enclosed within <answer> </answer> tags. i.e., <answer> 2 </answer> or <answer> 2,3,5 </answer>.
\end{tcolorbox}

\begin{tcolorbox}[colback=black!1!white,colframe=black!57!white,title=Prompt for Refiner Agent, breakable]

You are a caption refinement specialist in a caption-to-molecule generation system. Your task is to refine the input caption based on insights from the debate process.
\\
You will be provided with:\\
1. An original caption describing desired molecular properties\\
2. Multiple candidate molecules generated from this caption\\
3. Debates from multiple agents who evaluated these molecules and selected the most sufficient ones\\
Your goal is to analyze the debate and refine the original caption to make it more reliable for caption-to-molecule generation. Learn from the agents' reasoning to identify:\\
- What aspects of the caption led to successful molecule generation\\
- What ambiguities or unclear descriptions caused issues\\
- What key properties or constraints need to be emphasized or clarified\\
- What common mistakes can be avoided with better phrasing\\
\\
CRITICAL CONSTRAINTS:\\
- You MUST preserve the original meaning, purpose, and intent of the caption\\
- You MUST keep all original keywords and essential molecular properties\\
- You MUST NOT add new requirements that weren't in the original caption\\
- Only improve clarity, specificity, and reliability while maintaining the original scope\\
\\
Original caption: \{query\}\\
Candidate molecules generated: \{molecules\}\\
Agents' debate and selection: \{debate selections\}\\
\\
Based on the above information, please:\\
1. Analyze what aspects of the original caption were well-captured vs. what caused confusion\\
2. Refine the caption to improve its reliability for molecule generation\\
3. Ensure the refined caption maintains all original keywords, meaning, and purpose\\
Output your refined caption enclosed within <answer> </answer> tags.\\
Format: <answer> [Your refined caption here] </answer>

\end{tcolorbox}

\subsection{Details of Examiner Agent}
\label{sec:app_examiner_agent}

For the Examiner Agent, the set of chemical descriptors $\Pi$ includes MW: the average molecular weight of the molecule, LogP: the logarithm of the octanol-water partition coefficient value of the molecule, TPSA: the topological polar surface area of the molecule, HBD: the number of hydrogen bond donors in the molecule, HBA: the number of hydrogen bond acceptors in the molecule, RotB: the number of rotatable bonds in the molecule, AroRings: the number of aromatic rings for a molecule, Fsp3: the fraction of C atoms that are SP3 hybridized in the molecule, QED: the quantitative estimation of drug-likeness of the molecule, and MR: the molecular refractivity of the molecule.

To avoid evaluation leakage from the Examiner Agent, in the text-based open molecule generation task on S$^2$-Bench, the Examiner Agent does not report the LogP descriptor in the LogP subtask, the MR descriptor in the MR subtask, and the QED descriptor in the QED subtask.

An example of an examination report $\mathcal{E}$ generated by the Examiner Agent is

\begin{tcolorbox}[colback=black!1!white,colframe=black!57!white,title=Example of Examination Report, breakable]
The SMILES COC(=O)/C=C/[C@H]1\\CC[C@H]2[C@@H]3CCC4=CC(=O)\\CC[C@]4(C)[C@H]3C(=O)C[C@]12C is valid with the following properties:\\The average molecular weight of the molecule: 370.49. The logarithm of the octanol-water partition coefficient (LogP) value of the molecule: 4.04. The topological polar surface area (TPSA) of the molecule: 60.44. The number of hydrogen bond donors in the molecule: 0.00. The number of hydrogen bond acceptors in the molecule: 4.00. The number of rotatable bonds in the molecule: 2.00. The number of aromatic rings for a molecule: 0.00. The fraction of C atoms that are SP3 hybridized in the molecule: 0.70. The quantitative estimation of drug-likeness of the molecule: 0.54. The molecular refractivity (MR) of the molecule: 101.75.

\end{tcolorbox}

\subsection{More Experimental Results}
\label{sec:app_more_results}

\subsubsection{More Analysis of Multiple Perspectives}
\label{sec:app_more_analysis}

\noindent \textbf{Impact of developer-generalist LLMs debating}
We show the detailed ablation study results on the developer-generalist LLMs debating in Tables~\ref{tab:app_developer_generalist} and~\ref{tab:app_developer_generalist_tomg}.
In Table~\ref{tab:e-g_components} and Table~\ref{tab:app_developer_generalist},
D+D setting uses ChemDFM-R-14B, ChemDFM-v1.5-8B, and Chem-R-8B as LLMs in the Developer Agent, and ChemDFM-R-14B is used as the LLM in the Debate Agent and Refiner Agent.
In Table~\ref{tab:app_developer_generalist_tomg},
D+D setting uses ChemDFM-R-14B and Chem-R-8B as LLMs in the Developer Agent, and ChemDFM-R-14B is used as the LLM in the Debate Agent and Refiner Agent.
While G+G setting uses GPT-4o mini as the LLM in the Developer Agent, Debate Agent, and Refiner Agent in Tables~\ref{tab:e-g_components} and~\ref{tab:app_developer_generalist_tomg}.

\noindent \textbf{Impact of key components of Mol-Debate}
Table~\ref{tab:app_key_components_c2m} and Table~\ref{tab:app_key_components_tomg} show the detailed ablation study results on the key components of Mol-Debate on the caption-to-molecule generation task and text-based open molecule generation task, respectively.

\subsubsection{More Detailed Ablation Study}
\label{sec:app_more_ablation_results}

\noindent \textbf{Extensibility of the Mol-Debate on different backbone LLMs}
We show the ablation study of Mol-Debate using different backbone LLMs in all agents on the caption-to-molecule generation task in Tables~\ref{tab:app_diff_backbone} and~\ref{tab:app_diff_backbone_tomg}.
Our method surpasses the baseline LLMs, achieving the best performance on the majority of metrics.
This result demonstrates the extensibility of the proposed Mol-Debate framework across various backbone LLMs.

\noindent \textbf{Impact of different general backbone LLMs in Debate Agent and Refiner Agent of Mol-Debate.}
Tables~\ref{tab:llm_in_debate_agent_c2m} and~\ref{tab:llm_in_debate_agent_tomg} show that stronger general-purpose LLMs in the Debate and Refiner agents consistently improve Mol-Debate.
Replacing GPT-4o mini with GPT-5 mini yields higher EM and stronger similarity on caption-to-molecule generation, and substantially increases WSR across all text-based open molecule generation subtasks.
Further upgrading to GPT-5.2 brings only marginal gains on caption-to-molecule generation, suggesting diminishing returns once the debate backbone is sufficiently strong.

\noindent \textbf{Impact of hyperparameter of Mol-Debate}
We show the detailed ablation study results of Mol-Debate to three hyperparameters that control the number of debaters ($K$), debate rounds ($R$) and consensus score threshold ($\theta$), as shown in Table~\ref{tab:app_debater_number}, Table~\ref{tab:app_debate_rounds}, and Table~\ref{tab:app_threshold}, respectively.

\subsubsection{Samples of Generated Molecules}
\label{sec:app_samples}

\begin{figure*}[b]
    \centering
    \includegraphics[width=1\linewidth]{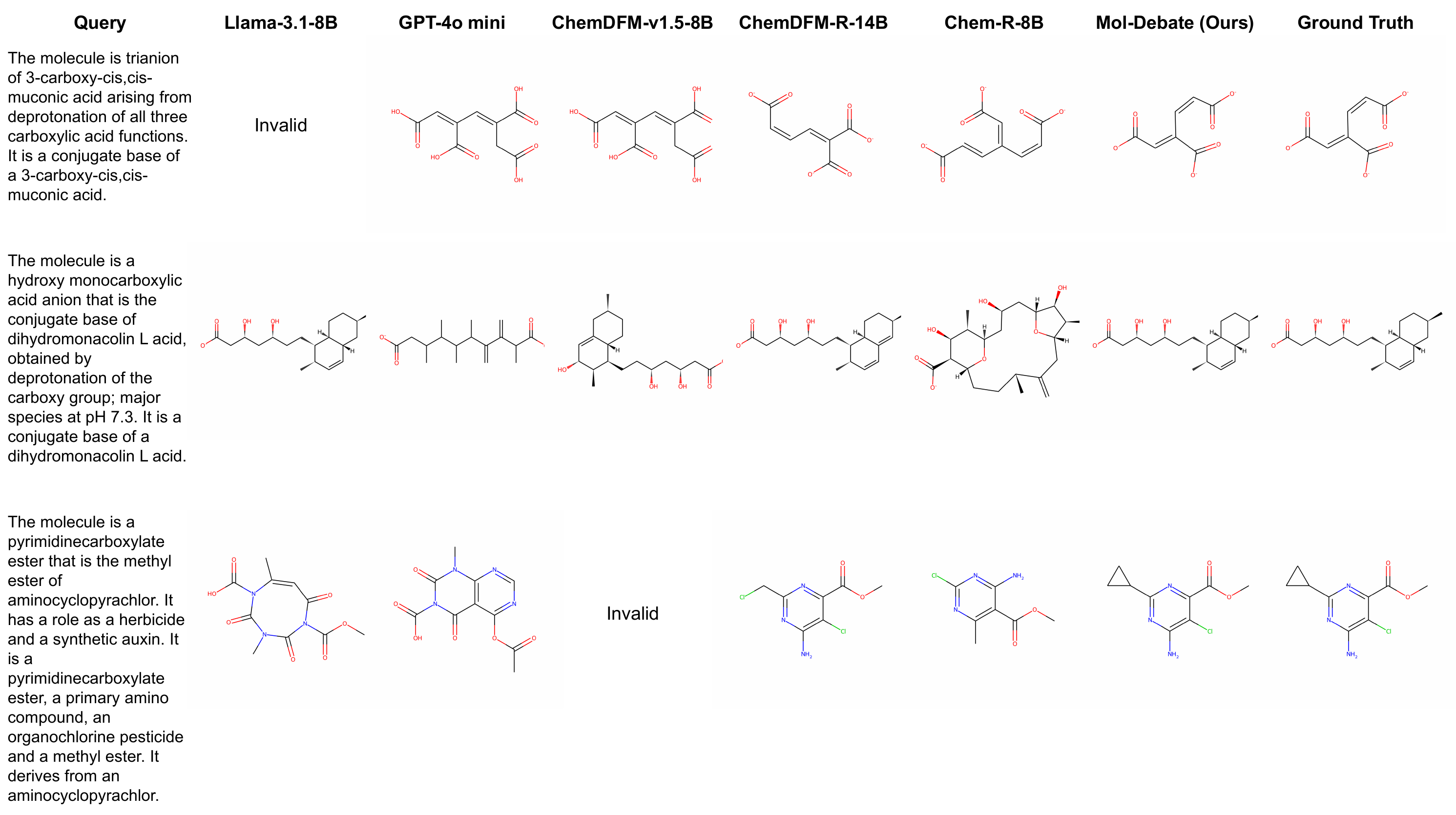}
    \caption{The generated samples of the caption-to-molecule generation task.}
    \label{fig:app_c2m_compare}
\end{figure*}

\begin{figure*}[b]
    \centering
    \includegraphics[width=1\linewidth]{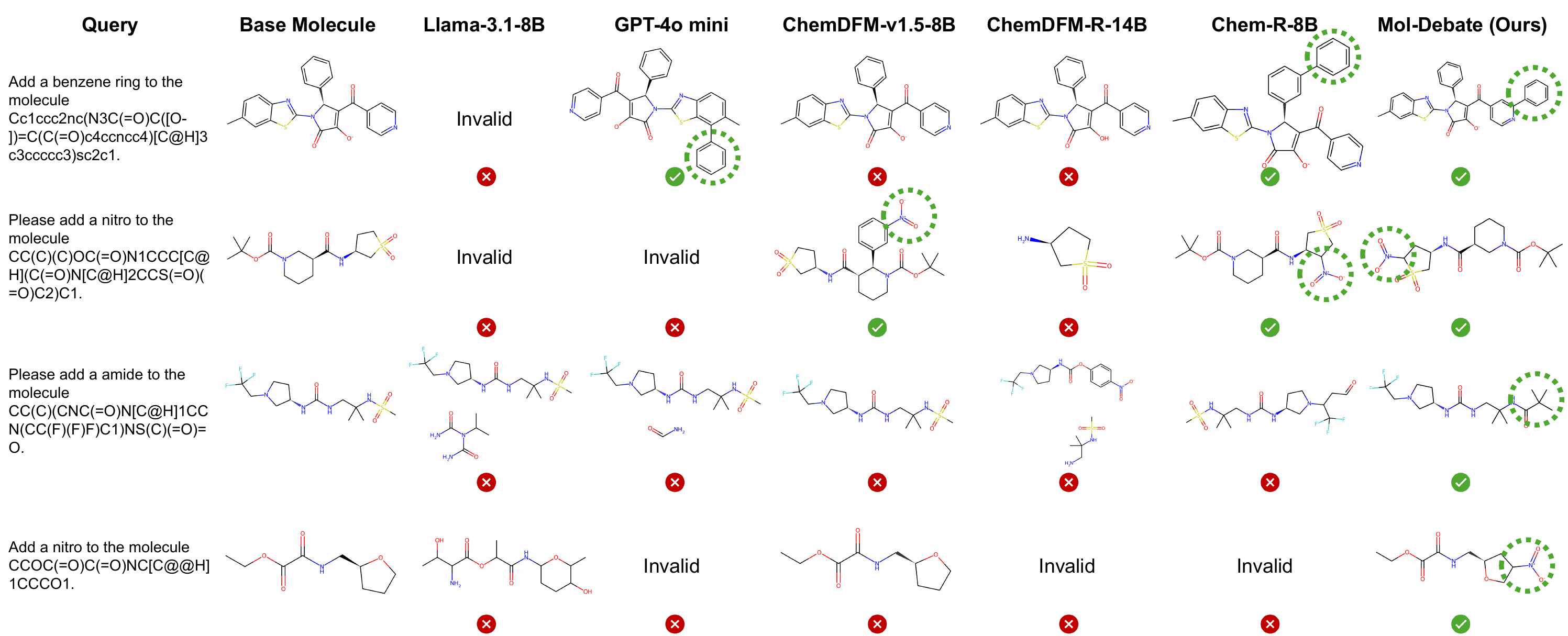}
    \caption{The generated samples of the text-based open molecule generation task.}
    \label{fig:app_tomg_compare}
\end{figure*}

We show the samples of generated molecules by Mol-Debate on the caption-to-molecule generation task in Figure~\ref{fig:app_c2m_compare} and text-based open molecule generation task in  Figure~\ref{fig:app_tomg_compare}.

\subsubsection{Case Study}
\label{sec:app_case}

\begin{figure*}[b]
    \centering
    \includegraphics[width=1\linewidth]{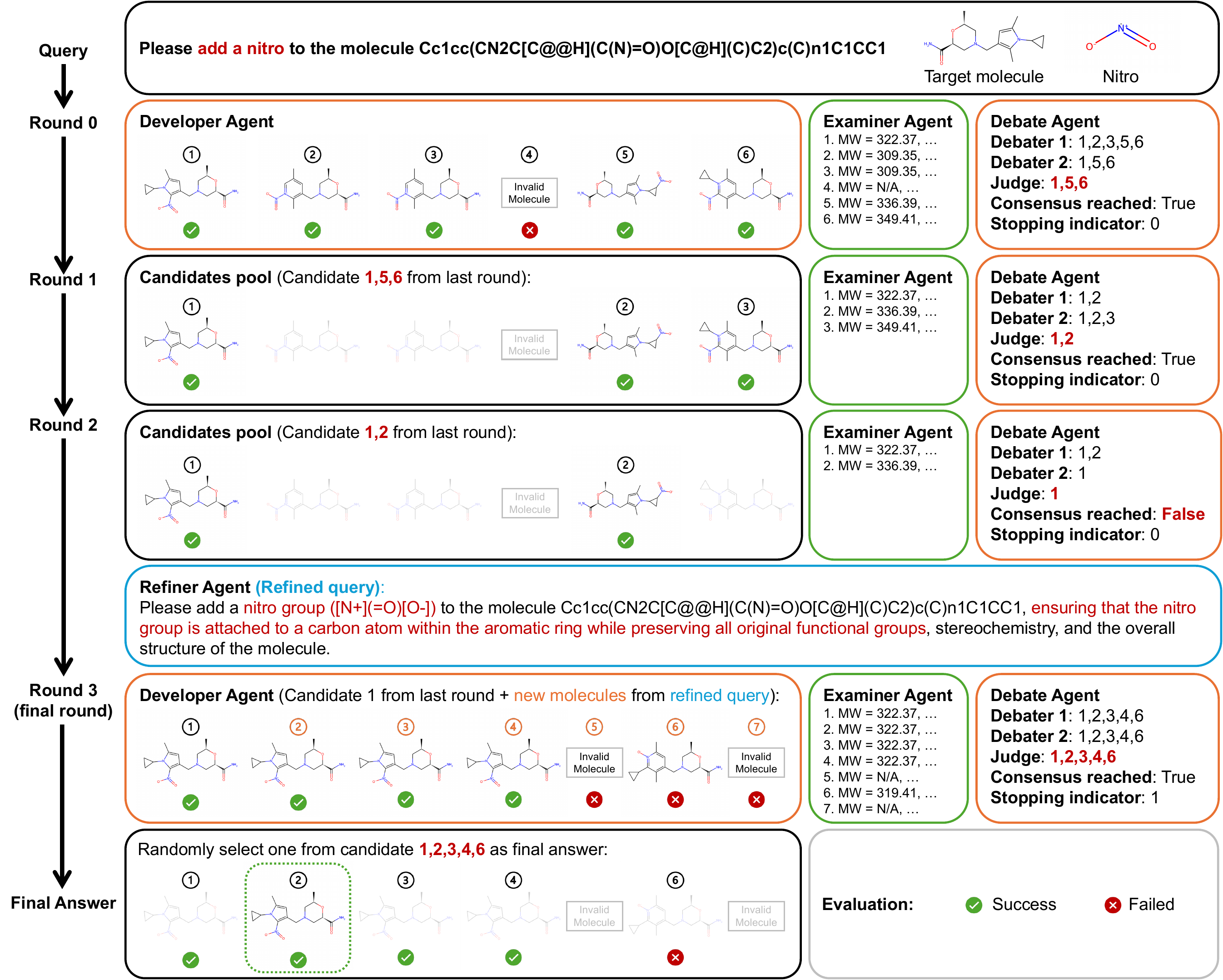}
    \caption{Case study of the Mol-Debate.
    In Round~0, the Developer Agent generates six candidates, and the Examiner Agent reports examination results, providing local structural evidence for the Debate Agent to form an initial candidate pool.
    Rounds~1 and~2 further narrow the pool, but the debaters disagree on the best nitro placement, triggering the Refiner Agent to rewrite the instruction with explicit constraints.
    Conditioned on the refined query, the final round expands the candidate set, and the Debate Agent converges to a consistent pool from which a valid, intent-satisfying molecule is selected as the final answer with a high success chance (4 out of 5).
    }
    \label{fig:case_2}
\end{figure*}

\begin{figure*}[b]
    \centering
    \includegraphics[width=1\linewidth]{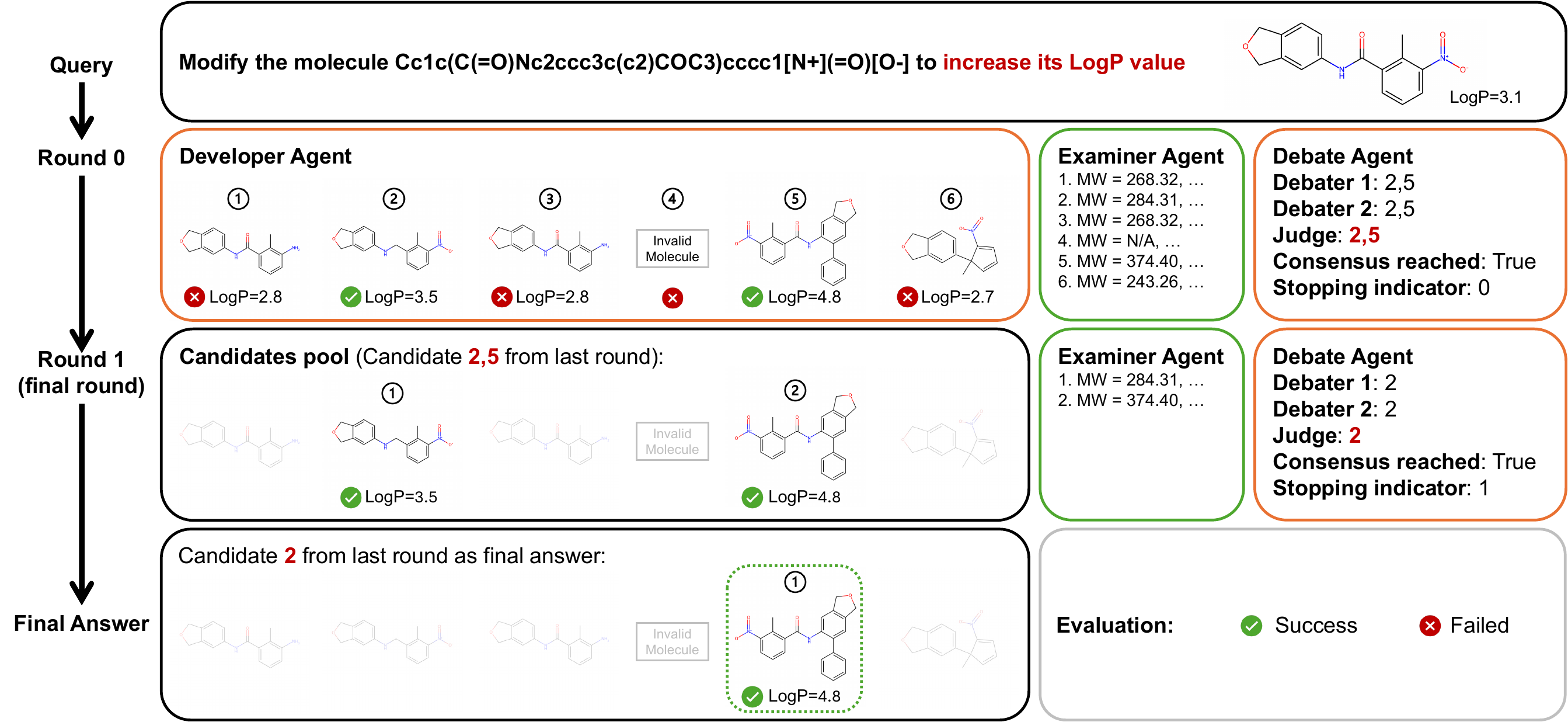}
    \caption{Case study of the Mol-Debate.
    In Round~0, the Developer Agent proposes six candidates and the Examiner Agent reports examination results, providing local evidence for intent-grounded selection. The Debate Agent retains the candidates that both remain valid and improve LogP (2 and 5).
    In the final round, the debate converges to a single candidate, and Mol-Debate outputs a successful molecule as the final answer.
    }
    \label{fig:case_3}
\end{figure*}

We show additional case studies of the Mol-Debate in Figure~\ref{fig:case_2} and Figure~\ref{fig:case_3}.

\subsection{Future Outlook of Mol-Debate Beyond Molecular Design}
\label{app:future_outlook}

The central idea of Mol-Debate is to use iterative multi-agent critique to progressively align semantic intent with structured outputs.
While this paper focuses on molecular design, this paradigm may generalize to a wider range of multimodal and scientific tasks.
Recent studies on multimodal reasoning,~\cite{wang2024exploring,jiang2024prior}, sign language translation,~\cite{gong2024llms,jiang2026think}, biomedical imaging~\cite{jiang2025self}, and image editing~\cite{shen2026competitioncoopetitioncoopetitivetrainingfree} indicate that many challenging tasks benefit from stronger intermediate reasoning signals and more reliable feedback mechanisms.
This suggests a broader future for Mol-Debate: extending debate-based refinement to tasks where correct prediction depends on repeatedly reconciling natural language instructions with structured, visual, or domain-constrained outputs.

\begin{table*}[http]
    \centering
    \adjustbox{max width=\linewidth}{
            \begin{tabular}{lcccccccccc}
            \toprule
             Settings & Generalist LLM & EM $\uparrow$ & BLEU $\uparrow$ &		Lev. $\downarrow$ &	MA.F $\uparrow$ &	RD.F $\uparrow$ &	MO.F $\uparrow$ &	FCD $\downarrow$ &	Val. $\uparrow$ \\
            \midrule
              \textbf{D+G (\textbf{Ours})} & GPT-4o mini  &	\colorbox{MyBlue}{\textbf{0.5542}} & \colorbox{MyBlue}{\textbf{0.7008}} &	\colorbox{MyBlue}{\textbf{31.1388}} &	\colorbox{MyBlue}{\textbf{0.9422}} &	\colorbox{MyBlue}{\textbf{0.8839}} &	\colorbox{MyBlue}{\textbf{0.8362}} &	\colorbox{MyBlue}{\textbf{0.1850}} &	\colorbox{MyBlue}{\textbf{1.0000}} \\
              
              D+D  & - & 	 \colorbox{MyLightBlue}{0.4942} &	\colorbox{MyLightBlue}{0.6996} & \colorbox{MyLightBlue}{33.2330} &	\colorbox{MyLightBlue}{0.9324} &	\colorbox{MyLightBlue}{0.8580} &	\colorbox{MyLightBlue}{0.8057} &	\colorbox{MyLightBlue}{0.2057} &	\colorbox{MyLightBlue}{0.9988} \\
              
             G+G  & GPT-4o mini &	0.0206 & 0.1188 & 228.6782 &	0.6033 &	0.3701 & 0.2825 &	6.1839 &	0.8791 \\
             \midrule
              \textbf{D+G (\textbf{Ours})} & GPT-5 mini  & \colorbox{MyBlue}{\textbf{0.5982}} &	\colorbox{MyBlue}{\textbf{0.7059}} &	\colorbox{MyBlue}{\textbf{30.8294}} &	\colorbox{MyBlue}{\textbf{0.9601}} &	\colorbox{MyBlue}{\textbf{0.9103}} &	\colorbox{MyBlue}{\textbf{0.8694}} &	\colorbox{MyBlue}{\textbf{0.1258}} &	\colorbox{MyLightBlue}{0.9973} \\
              
              D+D  & - & 	\colorbox{MyLightBlue}{0.4942} &	\colorbox{MyLightBlue}{0.6996} & \colorbox{MyLightBlue}{33.2330} &	\colorbox{MyLightBlue}{0.9324} &	\colorbox{MyLightBlue}{0.8580} &	\colorbox{MyLightBlue}{0.8057} &	\colorbox{MyLightBlue}{0.2057} &	\colorbox{MyBlue}{\textbf{0.9988}} \\
              
             G+G  & GPT-5 mini &	0.2300  &0.4605	 &	49.3379 &	0.8946 &	0.7467 &	0.6687 &	0.6228 &	0.9864 \\
             \bottomrule
        \end{tabular}
    }
    \caption{Detailed ablation study of Mol-Debate on the Developer-generalist LLMs Debating on the caption-to-molecule generation task. D: Developer LLM, G: Generalist LLM. \textcolorblue{\textbf{Blue}} marks the \textcolorblue{\textbf{best-performing}} method, \textcolorlblue{light blue} represents the \textcolorlblue{second-best-performing} method.}
    \label{tab:app_developer_generalist}
\end{table*}

\begin{table*}[t]
    \centering
    % \footnotesize
    \adjustbox{max width=\linewidth}{
    \begin{tabular}{lcc|cc|cc|cc}
        \toprule
        \multirow{2}{*}{Method} & \multicolumn{2}{c}{MolCustom} & \multicolumn{2}{c}{MolEdit} & \multicolumn{2}{c}{MolOpt}  & \multicolumn{2}{c}{\textbf{Average}}\\
        \cmidrule(lr){2-9}
          & SR $\uparrow$ & WSR $\uparrow$ & SR $\uparrow$ & WSR $\uparrow$ & SR $\uparrow$ & WSR $\uparrow$ & SR $\uparrow$ & WSR $\uparrow$\\
        \midrule
          \textbf{D+G (\textbf{Ours})} & \colorbox{MyBlue}{\textbf{0.2176}}  & \colorbox{MyBlue}{\textbf{0.1455}} &  \colorbox{MyBlue}{\textbf{0.8760}} & \colorbox{MyBlue}{\textbf{0.5923}} & \colorbox{MyBlue}{\textbf{0.9327}}  & \colorbox{MyBlue}{\textbf{0.6017}} &   \colorbox{MyBlue}{\textbf{0.6754}} &	\colorbox{MyBlue}{\textbf{0.4465}} \\
          D+D & 0.1493 &	0.0999 & 0.7462 &	0.5142 & 0.7527 &	\colorbox{MyLightBlue}{0.5085} & 0.5494 &	\colorbox{MyLightBlue}{0.3742}\\
          G+G & \colorbox{MyLightBlue}{0.1784} &	\colorbox{MyLightBlue}{0.1187} & \colorbox{MyLightBlue}{0.7913} &	\colorbox{MyLightBlue}{0.5153} & \colorbox{MyLightBlue}{0.8266} &	0.4213 & \colorbox{MyLightBlue}{0.5988} &	0.3518 \\

         \bottomrule
    \end{tabular}
    }
    \caption{Detailed ablation study of Mol-Debate on the Developer-generalist LLMs Debating on the text-based open molecule generation task. D: Developer LLM, G: Generalist LLM.}
        \label{tab:app_developer_generalist_tomg}
\end{table*}

\begin{table*}[http]
    \centering
    % \footnotesize
    \adjustbox{max width=\linewidth}{
    \begin{tabular}{lcccccccc}
        \toprule
         Methods & EM $\uparrow$ & BLEU $\uparrow$ &		Lev. $\downarrow$ &	MA.F $\uparrow$ &	RD.F $\uparrow$ &	MO.F $\uparrow$ &	FCD $\downarrow$ &	Val. $\uparrow$ \\
        \midrule
         
          \textbf{Mol-Debate (Ours)} &	\colorbox{MyBlue}{\textbf{0.5542}} & 0.7008 &	\colorbox{MyLightBlue}{31.1388} &	\colorbox{MyBlue}{\textbf{0.9422}} &	\colorbox{MyBlue}{\textbf{0.8839}} &	\colorbox{MyBlue}{\textbf{0.8362}} &	\colorbox{MyBlue}{\textbf{0.1850}} &	\colorbox{MyBlue}{\textbf{1.0000}} \\
            $-$ Examiner &	0.5030 & 	\colorbox{MyBlue}{\textbf{0.7238}} &	31.5545 &	0.9375 &	0.8737 &	0.8227 &	0.2139 &	0.9748 \\
            $-$ Refiner & \colorbox{MyLightBlue}{0.5324} & \colorbox{MyLightBlue}{0.7221} &	\colorbox{MyBlue}{\textbf{29.2915}} &	\colorbox{MyLightBlue}{0.9390} &	\colorbox{MyLightBlue}{0.8770} &	\colorbox{MyLightBlue}{0.8279} &	\colorbox{MyLightBlue}{0.1894} &	\colorbox{MyBlue}{\textbf{1.0000}} \\
            $-$ Examiner, $-$ Refiner &	0.5009 & 0.7190 &	32.1133 &	0.9356 &	0.8701 &	0.8185 &	0.2146 &	0.9730 \\
            
            \bottomrule
            
    \end{tabular}
    }
    \caption{Detailed ablation study for key components of Mol-Debate on the caption-to-molecule generation task. 
    }
    \label{tab:app_key_components_c2m}
    
\end{table*}

\begin{table*}[http]
    \centering
    % \footnotesize
    \adjustbox{max width=\linewidth}{
    \begin{tabular}{lcc|cc|cc|cc}
        \toprule
        \multirow{2}{*}{Method} & \multicolumn{2}{c}{MolCustom} & \multicolumn{2}{c}{MolEdit} & \multicolumn{2}{c}{MolOpt}  & \multicolumn{2}{c}{\textbf{Average}}\\
        \cmidrule(lr){2-9}
          & SR $\uparrow$ & WSR $\uparrow$ & SR $\uparrow$ & WSR $\uparrow$ & SR $\uparrow$ & WSR $\uparrow$ & SR $\uparrow$ & WSR $\uparrow$\\
        \midrule
          \textbf{Mol-Debate (Ours)} & \colorbox{MyBlue}{\textbf{0.2176}}  & \colorbox{MyBlue}{\textbf{0.1455}} &  \colorbox{MyBlue}{\textbf{0.8760}} & \colorbox{MyLightBlue}{0.5923} & \colorbox{MyLightBlue}{0.8619}  & \colorbox{MyLightBlue}{0.5650} &   \colorbox{MyBlue}{\textbf{0.6518}} &	\colorbox{MyLightBlue}{0.4342} \\
          
          $-$ Examiner & 0.1869 &	0.1244 & \colorbox{MyLightBlue}{0.8730} &	\colorbox{MyBlue}{\textbf{0.5929}} & 0.8244 & 0.5585 & 0.6281 &	0.4253\\
          
          $-$ Refiner & \colorbox{MyLightBlue}{0.2173} &	\colorbox{MyLightBlue}{0.1454} & 0.8595 &	0.5799 & \colorbox{MyBlue}{\textbf{0.8765}} &	\colorbox{MyBlue}{\textbf{0.5789}} & \colorbox{MyLightBlue}{0.6511} &	\colorbox{MyBlue}{\textbf{0.4347}} \\
          
          $-$ Examiner, $-$ Refiner & 0.1896 &	0.1263 & 0.8559 &	0.5798 & 0.8422 &	0.5706 & 0.6292 &	0.4256 \\

         \bottomrule
    \end{tabular}
    }
    \caption{Detailed ablation study for key components of Mol-Debate on text-based open molecule generation task.}
        \label{tab:app_key_components_tomg}
\end{table*}

\begin{table*}[http]
    \centering
    \adjustbox{max width=\linewidth}{
            \begin{tabular}{lccccccccc}
            \toprule
             Method & EM $\uparrow$ & BLEU $\uparrow$ &		Lev. $\downarrow$ &	MA.F $\uparrow$ &	RD.F $\uparrow$ &	MO.F $\uparrow$ &	FCD $\downarrow$ &	Val. $\uparrow$ \\
            \midrule
              GPT-4o mini & 0.0173 & \textbf{0.1947} &		\textbf{63.2667} &	0.5988 &	0.3629 &	0.3027 &	8.6415 &	0.5148 \\
              \quad\textbf{+ Mol-Debate} & \textbf{0.0206} & 0.1188 & 228.6782 &	\textbf{0.6033} &	\textbf{0.3701} & \textbf{0.2825} &	\textbf{6.1839} &	\textbf{0.8791} \\
                
                \midrule
                GPT-5 mini & 0.1773 & 0.4046 &	47.1945 &	0.8829 &	0.7280 &	0.6575 &	1.2423 &	0.7942 \\
                \quad\textbf{+ Mol-Debate} & \textbf{0.2300}  &\textbf{0.4605}	 &	\textbf{49.3379} &	\textbf{0.8946} &	\textbf{0.7467} &	\textbf{0.6687} &	\textbf{0.6228} &	\textbf{0.9864} \\

              \midrule
              ChemDFM-R-14B & 0.4948 & \textbf{0.6733} &		\textbf{33.9506} &	0.9313 &	0.8592 &	\textbf{0.8092} &	0.2236 &	0.9733 \\
              \quad\textbf{+ Mol-Debate} & \textbf{0.4994} &	0.6674 &	38.1255 &	\textbf{0.9336} &	\textbf{0.8612} &	0.8088 &	\textbf{0.2041} &	\textbf{0.9967} \\
            
            \midrule
            Gemma-4-31b-it &  0.0452 & \textbf{0.3786}	&	\textbf{49.8358} &	0.7067 &	0.4932 &	0.4169 &	3.8080 &	0.7061\\
            \quad\textbf{+ Mol-Debate} & \textbf{0.1694} & 0.2581 &		129.1752 &	\textbf{0.8459} &	\textbf{0.6738} &	\textbf{0.5985} &	\textbf{1.3463}	 & \textbf{0.9297} \\
              
             \bottomrule
        \end{tabular}
    }
    \caption{Ablation study of Mol-Debate using different backbone LLM on the caption-to-molecule generation task.}
    \label{tab:app_diff_backbone}
\end{table*}

\begin{table*}[http]
    \centering
    % \footnotesize
    \adjustbox{width=\linewidth}{
    \begin{tabular}{lcc|cc|cc|cc}
        \toprule
        \multirow{2}{*}{Method} & \multicolumn{2}{c}{MolCustom} & \multicolumn{2}{c}{MolEdit} & \multicolumn{2}{c}{MolOpt}  & \multicolumn{2}{c}{\textbf{Average}}\\
        \cmidrule(lr){2-9}
          & SR $\uparrow$ & WSR $\uparrow$ & SR $\uparrow$ & WSR $\uparrow$ & SR $\uparrow$ & WSR $\uparrow$ & SR $\uparrow$ & WSR $\uparrow$\\
      \midrule
          GPT-4o mini & 0.1063 & 0.0696 &  0.5059 & 0.3549 & 0.3574  & 0.2727 &   0.3232 & 0.2324 \\
          \quad\textbf{+ Mol-Debate} &  \textbf{0.1784} &	\textbf{0.1187} & \textbf{0.7913} &	\textbf{0.5153} & \textbf{0.8266} &	\textbf{0.4213} & \textbf{0.5988} &	\textbf{0.3518}  \\
        \midrule
          GPT-5 mini & 0.5164 &	0.3398 & 0.8865 &	0.6033 & 0.8454 &	0.4069 & 0.7494 &	0.4500\\
          \quad\textbf{+ Mol-Debate}  & \textbf{0.5937} &	\textbf{0.3959} & \textbf{0.9551} & \textbf{0.6545} & \textbf{0.9155} &	\textbf{0.4192} & \textbf{0.8214} & \textbf{0.4898} \\

         \bottomrule
    \end{tabular}
    }
    \caption{Ablation study of Mol-Debate with different backbone LLM on text-based open molecule generation task.}
        \label{tab:app_diff_backbone_tomg}
\end{table*}

\begin{table*}[http]
    \centering
    \adjustbox{max width=\linewidth}{
            \begin{tabular}{llccccccccc}
            \toprule
             Method & General LLM & EM $\uparrow$ & BLEU $\uparrow$ &		Lev. $\downarrow$ &	MA.F $\uparrow$ &	RD.F $\uparrow$ &	MO.F $\uparrow$ &	FCD $\downarrow$ &	Val. $\uparrow$ \\
             \midrule
             \textbf{Mol-Debate} & GPT-4o mini & 0.5542 & 0.7008 &	31.1388 &	0.9422 &	0.8839 &	0.8362 &	0.1850 &	\textbf{1.0000} \\
             \textbf{Mol-Debate} & GPT-5 mini & 0.5982 &	\textbf{0.7059} &	\textbf{30.8294} &	\textbf{0.9601} &	\textbf{0.9103} &	\textbf{0.8694} &	\textbf{0.1258} &	0.9973 \\
              \textbf{Mol-Debate} & GPT-5.2 & \textbf{0.6006} & 0.7008 &	31.3167 &	0.9587 &	0.9085 &	0.8663 &	0.1365 &	0.9979 \\
             \bottomrule
        \end{tabular}
    }
    \caption{Ablation study of Mol-Debate using the different general backbone LLMs in Debate Agent and Refiner Agent on the caption-to-molecule generation task.}
    \label{tab:llm_in_debate_agent_c2m}
\end{table*}

\begin{table*}[http]
    \centering
    % \footnotesize
    \adjustbox{width=\linewidth}{
    \begin{tabular}{llcc|cc|cc|cc}
        \toprule
        \multirow{2}{*}{Method} & \multirow{2}{*}{General LLM} & \multicolumn{2}{c}{MolCustom} & \multicolumn{2}{c}{MolEdit} & \multicolumn{2}{c}{MolOpt}  & \multicolumn{2}{c}{\textbf{Average}}\\
        \cmidrule(lr){3-10}
          & & SR $\uparrow$ & WSR $\uparrow$ & SR $\uparrow$ & WSR $\uparrow$ & SR $\uparrow$ & WSR $\uparrow$ & SR $\uparrow$ & WSR $\uparrow$\\
        \midrule
          Mol-Debate & GPT-4o mini & 0.2176  & 0.1455 &  0.8760 & 0.5923 & 0.8619  & 0.5650 &   0.6518 &	0.4342 \\
          Mol-Debate & GPT-5 mini & \textbf{0.3965} &	\textbf{0.2671} & \textbf{0.9399} &	\textbf{0.6432} & \textbf{0.9203} &	\textbf{0.6052} & \textbf{0.7522} & \textbf{0.5052} \\

         \bottomrule
    \end{tabular}
    }
    \caption{Ablation study of Mol-Debate using the different general backbone LLMs in Debate Agent and Refiner Agent on the text-based open molecule generation task.}
        \label{tab:llm_in_debate_agent_tomg}
\end{table*}

\begin{table*}[http]
    \centering
    \adjustbox{max width=\linewidth}{
    \begin{tabular}{ccccccccc}
        \toprule
         \#Debater ($K$) & EM $\uparrow$ & BLEU $\uparrow$ &		Lev. $\downarrow$ &	MA.F $\uparrow$ &	RD.F $\uparrow$ &	MO.F $\uparrow$ &	FCD $\downarrow$ &	Val. $\uparrow$ \\
        \midrule
           2  &	\colorbox{MyBlue}{\textbf{0.5542}} & 0.7008 &	31.1388 &	\colorbox{MyBlue}{\textbf{0.9422}} &	\colorbox{MyBlue}{\textbf{0.8839}} &	\colorbox{MyBlue}{\textbf{0.8362}} &	\colorbox{MyLightBlue}{0.1850} &	\colorbox{MyBlue}{\textbf{1.0000}} \\
           3  & 	\colorbox{MyLightBlue}{0.5485} & \colorbox{MyLightBlue}{0.7085} &	\colorbox{MyLightBlue}{30.2145} &	0.9400 &	\colorbox{MyLightBlue}{0.8811} &	\colorbox{MyLightBlue}{0.8351} &	0.1857 &	\colorbox{MyBlue}{\textbf{1.0000}} \\
          4  & 0.5373 & \colorbox{MyBlue}{\textbf{0.7157}} &		\colorbox{MyBlue}{\textbf{29.6958}} &	\colorbox{MyLightBlue}{0.9409} &	0.8810 &	0.8333 &	\colorbox{MyBlue}{\textbf{0.1807}} &	0.9997 \\
         \bottomrule
    \end{tabular}
    }
    \caption{Detailed ablation study of Mol-Debate on the number of debaters ($K$) on the caption-to-molecule generation task. \textcolorblue{\textbf{Blue}} marks the \textcolorblue{\textbf{best-performing}} method, \textcolorlblue{light blue} represents the \textcolorlblue{second-best-performing} method.}
    \label{tab:app_debater_number}
\end{table*}

\begin{table*}[http]
    \centering
    \adjustbox{max width=\linewidth}{
    \begin{tabular}{ccccccccc}
        \toprule
         \#Round ($R$) & EM $\uparrow$ & BLEU $\uparrow$ &		Lev. $\downarrow$ &	MA.F $\uparrow$ &	RD.F $\uparrow$ &	MO.F $\uparrow$ &	FCD $\downarrow$ &	Val. $\uparrow$ \\
        \midrule
         1 & 0.5358 & 0.7007 & 31.2652 & 0.9396 & 0.8773 & 0.8274 & \colorbox{MyBlue}{\textbf{0.1843}} & \colorbox{MyBlue}{\textbf{1.0000}} \\
         2  &  0.5433 & \colorbox{MyBlue}{\textbf{0.7039}} &		\colorbox{MyBlue}{\textbf{30.9088}} &	0.9410 &	0.8782 &	0.8280 &	0.1878 &	\colorbox{MyBlue}{\textbf{1.0000}} \\
        3 &  \colorbox{MyLightBlue}{0.5515} & 0.6978 &		31.4864 &	\colorbox{MyLightBlue}{0.9417} &	\colorbox{MyLightBlue}{0.8833} &	\colorbox{MyLightBlue}{0.8332} &	0.1894 &	\colorbox{MyBlue}{\textbf{1.0000}} \\
          4  &	\colorbox{MyBlue}{\textbf{0.5542}} & \colorbox{MyLightBlue}{0.7008} &	\colorbox{MyLightBlue}{31.1388} &	\colorbox{MyBlue}{\textbf{0.9422}} &	\colorbox{MyBlue}{\textbf{0.8839}} &	\colorbox{MyBlue}{\textbf{0.8362}} &	\colorbox{MyLightBlue}{0.1850} &	\colorbox{MyBlue}{\textbf{1.0000}} \\
         \bottomrule
    \end{tabular}
    }
    \caption{Detailed ablation study of Mol-Debate on debate rounds ($R$) on the caption-to-molecule generation task.}
    \label{tab:app_debate_rounds}
\end{table*}

\begin{table*}[http]
    \centering
    \adjustbox{max width=\linewidth}{
    \begin{tabular}{ccccccccc}
        \toprule
         Threshold ($\theta$) & EM $\uparrow$ & BLEU $\uparrow$ &		Lev. $\downarrow$ &	MA.F $\uparrow$ &	RD.F $\uparrow$ &	MO.F $\uparrow$ &	FCD $\downarrow$ &	Val. $\uparrow$ \\
        \midrule
          0.4  &	0.5361  & \colorbox{MyBlue}{\textbf{0.7063}}  &	\colorbox{MyBlue}{\textbf{29.9264}}  &	0.9392  &	0.8785  &	0.8285  &	\colorbox{MyLightBlue}{\textbf{0.1837}}  &	\colorbox{MyLightBlue}{\textbf{0.9997}} \\
          0.6  &	\colorbox{MyBlue}{\textbf{0.5542}} & 0.7008 &	31.1388 &	\colorbox{MyBlue}{\textbf{0.9422}} &	\colorbox{MyBlue}{\textbf{0.8839}} &	\colorbox{MyBlue}{\textbf{0.8362}} &	0.1850 &	\colorbox{MyBlue}{\textbf{1.0000}} \\
          0.8  &	\colorbox{MyLightBlue}{0.5485}  &	\colorbox{MyLightBlue}{0.7040}  &	\colorbox{MyLightBlue}{30.7024}  &	\colorbox{MyLightBlue}{0.9410}  &	\colorbox{MyLightBlue}{0.8819}  &	\colorbox{MyLightBlue}{0.8331}  &	\colorbox{MyBlue}{\textbf{0.1820}}  &	\colorbox{MyLightBlue}{0.9997} \\
         \bottomrule
    \end{tabular}
    }
    \caption{Detailed ablation study of Mol-Debate on the consensus score threshold ($\theta$) on the caption-to-molecule generation task.}
    \label{tab:app_threshold}
\end{table*}

\end{document}